\documentclass[10pt]{article} % For LaTeX2e
\usepackage[accepted]{tmlr}
\usepackage{amsmath,amsfonts,bm}

\def\eqref#1{equation~\ref{#1}}
\def\1{\bm{1}}

\DeclareMathAlphabet{\mathsfit}{\encodingdefault}{\sfdefault}{m}{sl}
\SetMathAlphabet{\mathsfit}{bold}{\encodingdefault}{\sfdefault}{bx}{n}

\DeclareMathOperator*{\argmin}{arg\,min}

\usepackage{url}

\usepackage{xspace}

\usepackage{hyperref}
\hypersetup{pagebackref,breaklinks,colorlinks}

\usepackage{multirow}

\usepackage{graphicx}

\usepackage{caption, subcaption}

\usepackage{float}

\usepackage[nolist,nohyperlinks]{acronym}

\usepackage{booktabs} 

\usepackage{tikz}

\begin{acronym}
\acro{tta}[TTA]{Test-Time Adaptation}
\acro{ttt}[TTT]{Test-Time Training}
\acro{sitta}[SITTA]{Single Image Test-Time Adaptation}
\acro{mae}[MAE]{Masked Autoencoders} 
\acro{sam}[SAM]{SegmentAnything Model}
\acro{ttaent}[Ent]{Entropy-Minimization}
\acro{ttaiou}[dIoU]{Deep-Intersection-over-Union}
\acro{ttaref}[Ref]{Mask Refinement}
\acro{ttaadv}[Adv]{Adversarial-Attack}
\acro{ttapl}[PL]{Pseudo-Labelling}
\acro{ttaaugco}[AugCo]{Augmentation-Consistency}
\acro{acdc}[ACDC]{Adverse Conditions Dataset with Correspondences}
\acro{iou}[IoU]{Intersection over Union}
\acro{miou}[mIoU]{mean Intersection over Union}
\acro{ce}[CE]{Cross-Entropy}
\acro{vit}[ViT]{Vision Transformer}
\acro{bce}[BCE]{Binary Cross Entropy}
\acro{fgsm}[FGSM]{Fast Gradient Sign Method}
\acro{pgd}[PGD]{Projected Gradient Descent}
\end{acronym}

\title{Single Image Test-Time Adaptation for Segmentation}

\author{\name Klara Janouskova
\email klara.janouskova@fel.cvut.cz \\
\addr Visual Recognition Group, Faculty of Electrical Engineering \\
Czech Technical University in Prague
\AND
\name Tamir Shor 
\email tamir.shor@campus.technion.ac.il\\
\addr Technion – Israel Institute of Technology, Haifa, Israel
\AND
\name Chaim Baskin
\email chaimbaskin@technion.ac.il \\
\addr Technion – Israel Institute of Technology, Haifa, Israel
\AND
\name Jiri Matas 
\email  matas@fel.cvut.cz \\
\addr Visual Recognition Group, Faculty of Electrical Engineering \\
Czech Technical University in Prague
}

\makeatletter
\DeclareRobustCommand\onedot{\futurelet\@let@token\@onedot}
\def\@onedot{\ifx\@let@token.\else.\null\fi\xspace}

\def\ie{\emph{i.e}\onedot}

\makeatother
\def\month{05}  % Insert correct month for camera-ready version
\def\year{2024} % Insert correct year for camera-ready version
\def\openreview{\url{https://openreview.net/forum?id=68LsWm2GuD}} % Insert correct link to OpenReview for camera-ready version

\begin{document}

\maketitle

\begin{abstract}

\ac{tta} methods improve domain shift robustness of deep neural networks. 
We explore the adaptation of segmentation models to a single unlabelled image with no other data available at test time. This allows individual sample performance analysis while excluding orthogonal factors such as weight restart strategies. We propose two new segmentation \ac{tta} methods and compare them to established baselines and recent state-of-the-art. The methods are first validated on synthetic domain shifts and then tested on real-world datasets. The analysis highlights that simple modifications such as the choice of the loss function can greatly improve the performance of standard baselines and that different methods and hyper-parameters are optimal for different kinds of domain shift, hindering the development of fully general methods applicable in situations where no prior knowledge about the domain shift is assumed.

Code and data: \url{https://klarajanouskova.github.io/sitta-seg/}

 % explain in more detail
       
\end{abstract}

\section{Introduction}
% TODO: We mention the IoU loss in contributions but it is nowhere in intro, where should we put it?
A common challenge in machine learning stems from the disparity between source (training) and target (deployment) data domains. Models optimized to minimize an error on a dataset from a specific domain are often expected to perform reliably in different domains.
The discrepancy between training and deployment data, known as the domain shift, 
is very common; in fact, few things do not change in time, and training happens (well) before deployment.
A domain shift may substantially degrade model performance at deployment time despite proper validation on training data, 
yet it is often not explicitly addressed and most machine learning effort has focused on the generalization problem. 

In many practical scenarios, the characteristics of the target domain are not known beforehand, making the preparation of the model with traditional domain adaptation \cite{rodriguez2019domain, tzeng2017adversarial} techniques nontrivial. 
Recent advances \cite{wang2020tent, sun2020test, gandelsman2022test} suggest that under certain weak assumptions about the domain shift - such as a stable label distribution across domains - it is possible to mitigate the performance degradation with methods based on the information carried by input data received at inference time.

\acf{tta} is suitable for a priori unknown or difficult-to-predict domain shifts. Characterized as an unsupervised and source-free technique, \ac{tta} adapts the model directly during inference. 
The source-free nature, \ie without access to the original training data, ensures compliance with data governance standards and enables adaptation in memory-constrained environments.

\acf{sitta} tailors a model at test time to each individual image. 
Since it operates on a single image, it does not introduce assumptions about the stability of the data distribution over time. Each time starting from the weights fixed at training time, \ac{sitta} is safe to use when any form of memorization of the deployment data is prohibited.  A disadvantage is an increased computational time compared to batch methods and the lack of retaining the acquired knowledge. On the other hand, \ac{sitta} could be leveraged to reduce the computational cost of adapting to a sequence of similar images by only processing the most informative samples.
Despite the advantages, only \cite{Khurana2021SITA:Adaptation} primarily addresses \ac{sitta}, the mainstream of \ac{tta} research deals with continual test-time adaptation in a changing environment \cite{niu2023towards, volpi2022road, wang2022continual}. 
These methods typically gradually update model parameters or accumulate image statistics for subsequent adaptation to individual images.
Continual \ac{tta} strategies are practical in many applications, such as autonomous driving, but they are challenging from the point of view of accurate assessment of relative strengths and weaknesses. The difficulty arises from the evaluations conducted on many images with varying levels and kinds of domain shift. 
Moreover, the sequence in which images are presented can significantly influence performance metrics, adding a layer of complexity to assessing the true efficacy of these methods. 
\ac{sitta} streamlines the evaluation process and contributes to understanding the broader class of \ac{tta} strategies. Continual \ac{tta} analysis has focused on complementary issues such as catastrophic forgetting.

 \begin{figure}[bt]
    \centering
        \includegraphics[keepaspectratio, width=\linewidth]{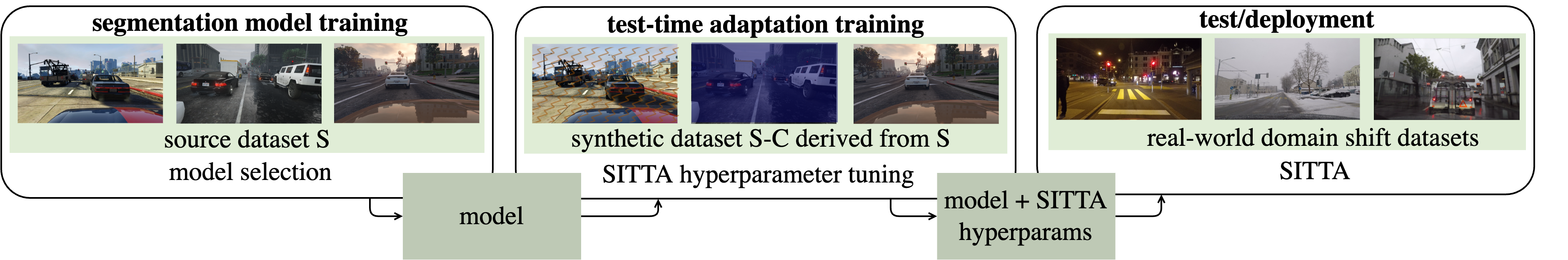}
    \caption{The proposed experimental framework for \acf{sitta}. Hyper-parameters are found on a synthetic dataset derived from the training set by applying a diverse set of corruptions. \ac{sitta} methods are then tested on real-world datasets with domain shift.}
    \label{fig:teaser}
\end{figure}

In this paper, we explore and improve the state of the art in \acl{sitta} for image semantic segmentation. While \ac{tta} has been applied to many tasks; the overwhelming majority of methods have been developed for image classification. Image segmentation \ac{tta} methods are each evaluated under very different conditions and compared to a limited set of baselines, making understanding their performance difficult.
We focus on semantic segmentation \ac{sitta} with self-supervised loss functions. These methods are broadly applicable across various segmentation models and domains without being constrained by specific architectural or segmentation training requirements.  
Consequently, the many works that rely on batch normalization layers are not included in this work as they are incompatible\footnote{At inference time, batch normalization uses per-feature mean and variance computed over the training dataset. \ac{tta} methods also incorporate the test data feature statistics. Transformer models typically use layer-normalization layers, which normalize features for each instance independently and thus always use the test sample statistics.} with the modern transformer architectures currently dominating the field. Likewise, methods such as image-reconstruction based \ac{tta} \cite{WangTest-TimeStreams, gandelsman2022test} are not included due to their significant demands for training process modification and model architecture adjustments. 
Some methods require access to training data before deployment since they train an auxiliary network. However, they do not alter the training of the segmentation model. 

Six \ac{tta} methods are evaluated in the \ac{sitta} setting.
Two are established baselines: entropy minimization \cite{Hu2019Depth-attentionalRemoval} and pseudolabelling  \cite{lee2013pseudo} (self-training). They are the only ones evaluated in the \ac{sitta} setup before. The third method \cite{prabhu2021augco} is a recently proposed segmentation \ac{tta} method that identifies confidently pseudolabelled pixels as consistent across augmentations. The fourth method \cite{nguyentipi} was proposed for image classification; we are the first to use it for image segmentation. The main idea is that if a network is robust to small adversarial domain shifts (it is optimized to not change the prediction under an adversarial attack at test time), it will also be robust to real-world domain shifts.

The remaining two approaches are novel. We combine and extend two techniques optimizing self-supervised losses with 1. learnt segmentation mask domain discriminators \cite{tzeng2017adversarial},
so far used for unsupervised domain adaptation and never applied to \ac{tta}, 
and 2. Segmentation mask refinement \footnote{Called ``denoising autoencoders'' in \cite{karani2021test}.} modules~\cite{karani2021test, valvano2021re},  never considered outside the medical domain \ac{tta}.

The idea of training-time unsupervised adaptation%
\footnote{In the domain adaptation literature, the term ``unsupervised'' refers to the situation when images from the target domain are available to the method, but they are not labelled.%
In such a setting, training a source-target domain discriminator is possible, which contradicts the standard implication of the term ``unsupervised''. The setting where no samples from the target domain are available during training is referred to as ``test-time adaptation'' or ``unsupervised source-free adaption''. } 
with mask discriminators
is that a classifier trained to distinguish masks from the source and target domains can be used to supervise adaptation. The source-domain segmentation model is finetuned to produce masks on the target domain that resemble those of the source domain, as predicted by the discriminator. This method is not directly applicable to \ac{tta} since the target images to train the discriminator are unavailable.
However, the target domain masks can be synthesized.
Copying random patches into different image 
locations was proposed in the medical domain. Such augmentation makes sense when the textures are more important than the shapes of objects, such as in the medical domain, but are not general enough for other images.
We propose to use the output of the segmenter on adversarially attacked images.
  % It is motivated by the intuition that the first pixels to be inverted in an untargeted adversarial attack are also those most likely to be impacted by domain shift.
  The adversarial optimization learning rate and the number of iterations control the severity of the domain shift. 
  Further, \cite{karani2021test} proposes to replace the mask discriminator with a denoising autoencoder. Since a discriminator can use shortcuts and only focus on the most discriminative parts of the masks, a model that produces a refined mask is trained to consider all the masks.  We explore both options.

We address the problem that arises from the practice in the current landscape of segmentation \ac{tta} that the performance assessment is carried out with inconsistent adaptation settings. 
For instance, keeping batch normalization statistics constant or updating them during entropy minimization can yield substantially different outcomes. If only one of the two options is tested \cite{Khurana2021SITA:Adaptation, volpi2022road}, contradictory results are reported.
Often, the evaluation compares only with the established baselines such as entropy minimization \cite{Wang2020Tent:Minimization} and batch normalization \cite{Ioffe2015BatchShift} statistics adaptation \cite{schneider2020improving, nado2020evaluating}, ignoring recent progress. 
 Many methods that improve these baselines have been proposed in recent years for both segmentation and classification \cite{NguyenTIPI:Invariance, Gao2023BackCorruption, niu2023towards, chen2022contrastive},  and their relative merit is unknown, since a comprehensive comparison with a well-defined methodology is lacking.

Our experimental framework, depicted in Figure \ref{fig:teaser}, consists of three stages: 1. \textbf{Segmentation model training}, which is standard. 2. \textbf{Test-time adaptation training} which tunes \ac{tta} hyper-parameters. We utilize an augmented version of the training datasets. This extension incorporates synthetic corruptions inspired by \cite{hendrycks2019robustness}. The corruption types include different kinds of noise and blur, weather conditions such as fog or frost, jpeg compression, and basic image intensity transformations. It can be derived from an arbitrary segmentation training dataset, as opposed to existing synthetic datasets used by previous work such as \cite{ros2016synthia}. It provides precise control over the conditions and facilitates detailed analysis. 3. \textbf{Evaluation} on real-world test datasets with domain shift. 

We experiment with two existing pretrained models for semantic segmentation. The models have different architectures and are trained on various datasets. Almost all segmentation \ac{tta} methods are evaluated on driving scenes benchmarks - we follow this practice and add a benchmark on common objects.

The main contributions of this paper are:
\begin{enumerate}
    \item We conduct a comparative study of six \ac{tta} techniques run in \ac{sitta} mode for image segmentation: Two established baselines, two adapted state-of-the-art methods from image classification and continual \ac{tta}, and two proposed methods. 
    \item Novel methods adapting ideas from unsupervised domain adaptation and medical imaging \ac{tta} to non-medical image segmentation are introduced, filling a gap in exploring diverse self-supervised loss functions. The method outperforms the other methods on multiple test datasets and is shown to be powerful on the images with the worst segmentation performance, as measured by \ac{iou}.
    % A novel adversarial refinement module training for \acl{ttaref} \acl{tta}.
    \item Improvements of baselines in the single-image setup by replacing \ac{ce} with the \ac{iou} loss.
    The performance of pseudo-labelling is improved by 3.51 \% and 3.28 \% on GTA5-C and COCO-C validation sets while with \ac{ce} loss, the improvements are by 1.7 \%  and 2.16 \% only, respectively.
    \item The first work shows the potential of \ac{sitta} for segmentation, an underexplored setup essential in applications with strict data governance standards or high variability among individual images.
    % \item The analysis of \ac{sitta} performance. The results reveal the unpredictability and variability of target data domains compared to source domains.
    \end{enumerate}

\section{Background}

% on TTA as special case of domain adaptation, related tasks 

Common approaches to domain adaptation change the style of labelled source images to resemble the training images \cite{tzeng2017adversarial, zhang2018task} or train domain classifiers to guide the adaptation process. In practice, this is not always feasible since source data may not be available for example for privacy or memory limitation reasons, or we may only have a small number of target domain images available when data arrive individually/in small batches, rather than all at once. In continually evolving environments, the distribution may change by the time adaptation on a large target dataset is completed. Various modifications of the traditional domain adaptation scenario tackling the aforementioned limitations have recently emerged, for example by considering no access to source data or a continual domain shift \cite{liu2021source, volpi2022road, wang2022continual, bartler2022mt3}. 

In particular, test-time adaptation methods assume no source data is available and aim to exploit the information from as little as a single target domain image. Like other domain adaptation methods, \ac{tta} methods are often inspired by semi-supervised learning methods. For instance, the most common \ac{tta} baseline relies on minimization of the predictions entropy, a method inspired by \cite{Saito2019Semi-supervisedEntropy}. Other methods rely on adapting the batch normalization statistics, inspired by methods like adaptive batch normalization \cite{li2018adaptive}, or aggregating statistics to create so-called prototypes \cite{tanwisuth2021prototype} that can be used to build a classifier.
% \cite{rodriguez2019domain, saito2019semi}

Some works also distinguish between \ac{tta} and \ac{ttt}. The difference between \ac{tta} and \ac{ttt} is that \ac{tta} methods such as \cite{nguyentipi, karani2021test} can be applied to arbitrary pre-trained models without any additional constraints while \ac{ttt} methods 
like \cite{gandelsman2022test, bartler2022mt3, liu2021ttt++} require modifications to the training process. However, not all works make this distinction and the boundary is not always clear, as some methods like \cite{karani2021test} need to train an auxiliary deep net on the source data but do not modify the model pretrained weights. In this work, both will be jointly referred to as \ac{tta} for simplicity.

In Appendix \ref{sec:ap:rel}, other related domain adaptation scenarios and their relation to \ac{tta} are described.

Generally, \ac{tta} methods can be split into three groups: Adaptation in the input space, feature space and output space. \textbf{Input space adaptation} aims to translate the images from the source domain to the input domain. In practice, \cite{Gao2023BackCorruption} achieve this by feeding target images with added noise to a diffusion model trained on the source data, coupled with reconstruction guidance to preserve semantics. The model doesn't retain any knowledge from the adaptation - the advantage is it is not susceptible to catastrophic forgetting but the disadvantage is it may limit the adaptation capabilities.
\textbf{Adaptation in the feature space} is the most common approach and typically relies on optimizing the network parameters via a self-supervised loss function. This can be done directly, i.e. through prediction entropy minimization, or by training an auxiliary task such as image reconstruction. Another set of feature-adaptation approaches are parameter-free and rely on accumulating the image statistics, such as the mean and variance of image features, or by aggregating confident prediction features into so-called prototypes, which are then used for classification.
\textbf{Output space adaptation} techniques aim to improve the network output without neither altering the network parameters and statistics nor the input image. This is done for instance in \cite{karani2021test} where an auxiliary network is trained to predict a refined mask. To the best of our knowledge, output space adaptation methods are typically only used to provide pseudo-masks, turning them into feature-space adaptation methods. This helps to iteratively improve the pseudo-masks and adapt to larger domain shifts.
All the methods evaluated in this work can be considered as feature space adaptation methods, possibly via output space adaptation.

% All of the methods can be considered as adapting to input $x$ by minimizing a self-supervised loss function $\mathcal{L}_{\text{TTA}}$ over a set of parameters $\omega \in \theta $ of the segmentation network $f_{\theta}$: $ \omega^\star = \argmin_{\omega}  \mathcal{L}_{\text{TTA}}(f, x)$. 

% The implementation of $\mathcal{L}_{\text{TTA}}(f, x)$ is what distinguishes the different methods. There are many possible choices for $\omega$ and the right choice may depend on time constraints, available GPU memory and the amount of data.
% In this work, two options are considered: $\omega = \theta $ where all the learnable parameters are optimized and $\omega = \theta_n $ where only the learnable parameters of the normalization layers such as BatchNorm, GroupNorm or LayerNorm are optimized. Details of $\mathcal{L}_{\text{TTA}}(f, x)$ implementation for the baseline methods and the necessary adaptations to the single-image setup are provided in Appendix \ref{sec:ap:baselines}.

\section{Related Work}

\textbf{Test-Time Adaptation methods for classification.}
Many recent methods propose improved strategies to update the batch normalization statistics \cite{schneider2020improving, nado2020evaluating}. A limitation of these methods is the reliance on presence of batch nromalization, which is often not part of recent transformer-based architectures.
In \cite{wang2020tent}, the learnable parameters of the normalization layers are also updated via entropy minimization. While this method is often reported as unstable since single-image statistics may not be sufficient, the method can also only update the normalization layers learnable parameters, without the statistics update, making it generalizable to all currently used architectures. 

 On classification tasks, many methods outperforming the aforementioned baselines have been proposed. A combination of self-supervised contrastive learning to refine the features and online label refinement with a memory bank is proposed in \cite{chen2022contrastive}. Recently, a method based on updating the parameters of the normalization layers of the network by optimizing it for robustness against adversarial perturbation as a representative of domain shift was proposed in \cite{nguyentipi}, outperforming similar test-time adaptation approaches.
 Rotation prediction is proposed in \cite{sun2020test} as self-supervised task to be learnt alongside the main one and then optimized at inference time. Lately, it was shown that reconstruction with masked auto-encoders is a very strong self-supervised task for test-time adaptation of classifiers by \cite{gandelsman2022test}.

\textbf{Test-Time Adaptation methods for segmentation.}
To the best of our knowledge, the only work also focused on adaptation to a single isolated image \cite{Khurana2021SITA:Adaptation} is based on computing the statistics from augmented version of the input image, assuming batch normalization layers are present in the network. 
Both \cite{prabhu2021augco} and \cite{wang2022continual} exploit augmented views of the input images to identify reliable predictions.
The method of \cite{prabhu2021augco} is based on the consistency of predictions between augmented views, which replaces prediction confidence for selecting reliable pixels. Cross entropy loss is then minimized on such reliable predictions, together with a regularization based on information entropy \cite{li2020rethinking} to prevent trivial solutions. The method achieves impressive results, however, in contrast to our experiments, knowledge of the target domain shift is used for hyper-parameter tuning.
The evaluation assumed a full test set available at once, focusing on source-free domain adaptation, rather than \ac{tta}, but the method is applicable to the \ac{tta} setup as well. 
In \cite{volpi2022road}, the performance of entropy minimization in a continual setup is explored, proposing parameter restart to tackle weight drift, significantly improving performance. The focus is on driving datasets only. Similarly, \cite{wang2022continual} also focus on continual adaptation. Again, augmentations of the images are generated to obtain more reliable predictions. Further, the network parameters are stochastically reset to their initial values to prevent forgetting of the source domain knowledge.

\textbf{Test-Time Adaptation methods for medical imaging.}
In \cite{karani2021test}, an autoencoder is proposed that translates predicted masks into refined mask. At test time, the segmenter is optimized to produce masks closer to the enhanced ones. However, this work assumes the whole test dataset is available at once, in contrast to our single-image setup.
The work of \cite{valvano2021re} is similar to \cite{karani2021test} but instead of a masked-autoencoder, a GAN-like discriminator trained end-to-end together with the segmenter is used, as well as an auxiliary reconstruction loss.

These works assume domain shifts specific to the medical imaging domain such as the use of a different scanner and thus make the assumption that only low-level features are affected. Under this assumption, these works typically optimize a small adapter only, ie. the first few convolutional layers of the segmenter. Nonetheless, these methods are generalizable to image segmentation.

\textbf{Enhancing existing \ac{tta} benchmarks.} There are multiple concurrent works that identify similar issues and reporting results consistent with our experiments, mostly for image classification. 
The work of \cite{yu2023benchmarking} also highlights the issue of evaluating each method under very different conditions and provides a benchmark for image classification \ac{tta} encompassing different adaptation scenarions, as well as diverse backbones and domain shift datasets. Similarly to ours, a significant disparity between synthetic corruptions performance and natural shifts is observed. However, the hyper-parameters were selected based on a single kind of domain shift, which may bias the results.
Another work adressing the issue of fair comparison of \ac{tta} methods is that of \cite{mounsaveng2024bag} which provides an analysis of existing orthogonal classification \ac{tta} methods. Class rebalancing is one of the tricks proposed to improve the methods' performance. Also, sample filtration to remove noisy high-entropy images is employed. In contrast, we analyze performance of different methods based on prediction entropy, showcasing some methods can actually be highly effective on those noisy, high-entropy samples. Similarly to ours, the work shows that baselines can be greatly improved by very simple tricks. 
In \cite{niu2023towards}, label imbalance at test-time is again identified as an important factor harming the \ac{tta} performance. Again, the works focus is on image classification and epxlores different normalization layer kinds and stabilization techniques of entropy minimization while we focus on comparions of cross-entropy and a class-imbalance aware segmentation loss function, the \ac{iou}.
Finally, \cite{yi2023critical} study \ac{tta} for image segmentations and how well classification methods transfer to semantic segmentation \ac{tta}. They conclude that many of the classification \ac{tta} improvements do not transfer to segmentation and again highlight the class imbalance, which is typically greater for segmentation datasets.

\section{Methods} \label{sec:method}

In total, six different methods are implemented and evaluated, including traditional \ac{tta} baselines, methods from other tasks and novel methods. All the methods consist of optimizing a self-supervised loss, the specifics of the loss being what differentiates the methods.
It can be formalized as follows:

\begin{equation*}
\theta_{i+1} = \arg\min_{\theta_i} \mathcal{L}(f_S^{\theta_i}, x)
\end{equation*}

where $x$ is the input image, $\theta_i$ are the parameters of the segmentation network $f_S^{\theta_i}$at the $i$-th iteration and $\mathcal{L}$ is the self-supervised loss function.

The methods considered are:
\begin{itemize}
    \item \textbf{\acf{ttaent}}, a method proposed by \cite{wang2020tent} inspired by semi-supervised learning where the self-supervised objective is the prediction entropy. It has been used as a baseline by most of the \ac{tta} work. Only normalization layer parameters are typically updated to reduce the computational cost. Whether batch normalization statistics are updated or not varies.
    
    \item\textbf{\acf{ttapl}}, also commonly referred to as self-training. The model is finetuned with pseudo-labels obtained from the pretrained segmentation model. There are many improvements and modifications. The standard approach is to threshold the predicted probabilities and only train the model on the most confident predictions.

    \item \textbf{\acf{ttaaugco}}, proposed by \cite{prabhu2021augco}, is a method based on self-training enhanced by also optimizing for consistency between the original prediction and the prediction on augmented views, adapted to the single isolated image scenario.

    \item \textbf{\acf{ttaadv}} is the method proposed by \cite{nguyentipi} for image classification \ac{tta}, adapted to the single, isolated image segmentation.

    \item \textbf{\acf{ttaref}} is one of the proposed methods and can be considered an enhanced pseudolabelling method. The pseudo-labels are obtained by a learnt refinement module that takes logit masks as inputs and outputs a refined segmentation mask. A method based on this idea has been implemented in medical imaging \cite{karani2021test} but never tested on non-medical tasks.

    \item \textbf{\acf{ttaiou}}, the second proposed method is similar to \ac{ttaref}. However, a single-scalar quality estimate is predicted by a learnt module and minimized at test time. It is similar to using a GAN-like discriminator, as done in unsupervised domain adaptation literature \cite{tzeng2017adversarial}.
\end{itemize}

Only the necessary modifications to make the methods applicable to the single, isolated image segmentation setup with no assumptions of specific network architecture were applied to the existing methods.

The proposed methods based on learnt mask-refinement and mask-discriminator modules will be described in the rest of this section.
A description of other methods and the details of their modifications in this work are in Appendix \ref{sec:ap:baselines}. 

The method proposed in this work is presented first and the differences from previous work are detailed aftwerwards.

\textbf{\ac{tta} with mask refinement} is based on the idea that since the output space changes much less than the input space, a mask translation module can be learnt to refine mask predictions on images from target distribution to resemble the masks obtained from source images. 
At test time, the refinement network can be viewed as an enhanced pseudo-label generation method. These pseudo-labels can then be used both as supervision for the segmenter or a direct replacement of the segmentation output without any parameter optimization. However, the second option is unlikely to tackle highly distorted masks since the refined mask cannot improve gradually.

To train the refinement network $f_\text{R}^{\phi}$ with learnable parameters $\phi$, images from the source distribution and the pretrained segmentation network $f^{\theta}_\text{S}$ are required. Given an image $x$ and $x'$ generated from $x$ by synthesizing a covariate domain shift (not changing the label), let us denote as $s = f^{\theta}_\text{S}(x)$ and $s' = f^{\theta}_\text{S}(x')$ the corresponding segmentation masks.
Then, $f_\text{R}$ is trained to predict $s$, given $s'$ as input:

\begin{equation}
    \argmin_{\phi} \mathcal{L}_{\text{CE}}(f_\text{R}^{\phi}(s'), s)
\end{equation}

Predicted masks $s$ can also be replaced with ground truth $g$ at training time:

\begin{equation}
    \argmin_{\phi} \mathcal{L}_{\text{CE}}(f_\text{R}^{\phi}(s'), g)
\end{equation}

where $\mathcal{L}_{\text{CE}}$ is the cross-entropy loss.

At test-time, adapting to an image $x$, the model parameters are updated to minimize the \ac{iou} loss between mask prediction and a refined mask estimated by $f_\text{R}$:
% \theta_i^\star
\begin{equation}
    \theta_{i+1} = \argmin_{\theta_i} \mathcal{L}_{\text{IoU}} (f_\text{R}( \overline{f}^{\theta_i}_\text{S}(x)), f^{\theta_i}_\text{S}(x))
\end{equation}

where $\theta_i$ are the learnable parameters of $f^{\theta}_\text{S}$ at optimization iteration $i$ and $\overline{f}_S$ denotes no gradient flow throughout the computations of $f_\text{S}$. 

\begin{figure}[tb]
    \centering
        \includegraphics[keepaspectratio, width=\linewidth]{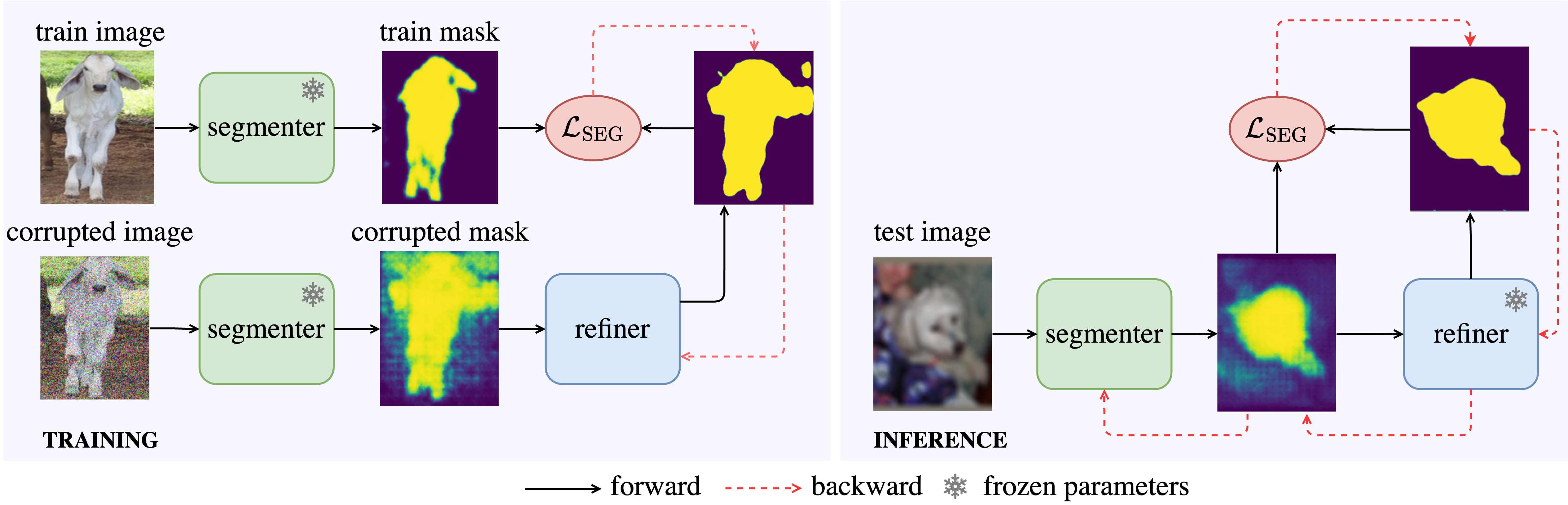}
    \caption{Mask refiner training (left) and \acf{ttaref} \ac{tta} inference (right). During training, the segmenter outputs masks from a training image and a corrupted version of the training image simulating domain shift. The mask refiner is then trained to predict the clean image mask given the corrupted image mask as input only - no gradients flow back to the segmenter. At inference time, the segmenter output is fed into the refiner model. The refined output is then used as a pseudo-label to finetune the segmenter. A single gradient update is performed in each \ac{tta} iteration, then the masks are updated. The segmenter output may change with the updated weights, which in turn results in a new, possibly better, pseudo-label from the refiner. Visualized on single class prediction.}
    \label{fig:refinement_train}
\end{figure}

An overview of the training pipeline and the \ac{tta} with mask-refinement is in Figure \ref{fig:refinement_train}.

\textbf{Refinement module training} requires generating masks resembling those that the model would output under domain shift. Since \ac{tta} assumes the domain shift is not known in advance, the goal is to synthetically generate diverse realistic segmentation masks representing masks predicted on images with domain shift. The advantage of the refinement module is that only the output space corrupted masks are needed. It doesn't matter how these were obtained since the refinement module is independent of the input images. This work simulates the mask corruptions by using mask predictions on the images from the first few iterations of a \ac{pgd} \cite{madry2017towards, kurakin2018adversarial} adversarial attack, using the inverted mask as target. The more iterations of the attack, the higher the mask corruption, but the less realistic it becomes. Examples of generated corrupted masks are shown in Appendix \ref{sec:ap:adv_ref}. The intuition behind this adversarial approach is that in the first iterations, the most challenging pixels for the network are converted. Similarly, those image areas could be easily impacted by domain shift. The idea of using adversarial attack as a representative of domain shifts was also used in \cite{nguyentipi} and it was previously shown that adversarial robustness improves robustness to some domain shift kinds \cite{croce2020robustbench}.

\textbf{\ac{tta} with Deep-Intersection-over-Union} is almos equivalent to \acf{ttaref}. The only difference is that rather than learning a refined mask, the network is optimized to predict the value of the \ac{iou} loss. The \ac{iou} loss can be computed from ground-truth at training time but the network receives the predicted mask as input only. 

\textbf{Difference from unsupervised domain adaptation literature:}
There are many domain adaptation methods based on learning a discriminator between source and target domain. The key difference is that in this work, a module estimating how corrupted the mask is or a mask refinement module is learnt, instead of a binary classifier. Hence, the output is either a new segmentation mask, or a real number, rather than a class (domain) probability. This has the advantage that the mask needs to be considered as a whole, rather than allowing the network to focus on the most discriminative parts only. Also, the target domain is not known in advance and cannot be used to train the refinement network - masks resembling the target domain masks need to be synthesized. 

\textbf{Difference from medical image segmentation \ac{tta} with denoising autoencoders}
The idea of learning to predict a refined mask in the output space is the same as in in \cite{karani2021test}. The main differences stem from the domain difference, since their approach is tailored to the medical domain. In \cite{karani2021test}, the corrupted masks are obtained through swapping input image patches, a heuristic method with many hyper-parameters. Such augmentation makes sense when texture is more important than shapes, which is often not the case outside the medical domain. This work proposes  exploiting adversarially-attacked images to synthesize corrupted masks.
Further, this work excludes methods that make changes to the segmentation model architecture while \cite{karani2021test} rely on an additional normalization module. Their module architecture is designed with domain shifts such as intensity transformations, typical for the medical domain, in mind. Updates of parameters in the segmentation network itself are not allowed during \ac{tta}.

\section{Experiments} \label{sec:experiments}

The structure of this section is as follows:

\begin{enumerate}
% \begin{itemize}
\item  \textbf{Evaluation metrics:} Given the focus of this study on \ac{sitta} and per-image performance analysis, we underscore the need for an image-level evaluation metric. The widely used \acf{miou} metric is typically applied at the dataset level and its adaptation for image-level assessment is not standardized.
\item  \textbf{Experimental Setup:} Experiment settings shared across experiments such as network architectures or hyper-parameters. Creation of the synthetic \ac{sitta} training set derived from the segmentation training dataset is also explained.
\item \textbf{Experimental Results and Analysis:} Experiment results and analysis.  The \ac{tta} methods are evaluated on two semantic segmentation models pretrained on the GTA5 \cite{richter2016playing} and COCO \cite{lin2014microsoft} datasets. 
% \end{itemize}
\end{enumerate}

\subsection{Evaluation metrics}
The standard semantic segmentation evaluation metric is the \ac{miou}, where the \ac{iou} score of each class is computed from predictions aggregated over the whole dataset and the per-class scores are averaged
\begin{equation}
    \text{mIoU} = \frac{1}{\text{C}} \sum_{k=1}^{\text{C}} \text{IoU}_k (m_{k}, g_{k})
\end{equation}
where $m_k, g_k$ are the predictions and ground truth values for class $k$ for all pixels across all images. Concatenating all the masks into a single one and then computing the metric would not change the results, each pixel has the same weight independent of the image size or difficulty.
This metric does not consider the size of objects or the difficulty of individual images. Per-image results cannot be compared, since not all classes are typically present in an image an it is not clear what value the score for that class should be. 

Two additional metrics are introduced to account for the limitations of the standard mIoU and make the evaluation more fine-grained. The first metric is designed to consider class imbalance and difficulty of individual images, focusing on per-class performance. It will be referred to as  
$\text{m}\overline{\text{IoU}}_c$ and is defined as

\begin{equation}
    \text{m}\overline{\text{IoU}}_c = \frac{1}{\text{C}} \sum_{k=1}^{\text{C}} \frac{1}{|\text{I}_k|} \sum_{i \in \text{I}_k} \text{IoU} (m_{ik}, g_{ik}) 
\end{equation}

where $\text{I}_k$ is the set of images in which either the prediction or the ground truth mask contains class $k$ and $|\text{I}_k|$ is the total number of images in $\text{I}_k$. 

 The second metric is focused more on per-image performance and can be computed for a single image. It will be referred to as $\text{m}\overline{\text{IoU}}_i$ and is defined as

\begin{equation}
    \text{m}\overline{\text{IoU}}_i = \frac{1}{|\text{I}|} \sum_{n \in \text{I}} \frac{1}{|\text{C}_n|} \sum_{k \in \text{C}_n} \text{IoU} (m_{nk}, g_{nk}) 
\end{equation}

where $\text{C}_n$ is the set of classes in the predicted masks or the ground truth of image $n$. $I$ denotes the set of all images. This is the metric reported in our experiments unless stated otherwise. It allows for per-image performance comparison with the disadvantage of not accounting for class imbalance - less frequent classes (on the image level) get smaller weight.

Similar metrics were recently considered by other works \cite{volpi2022road}, typically only aggregating over images where the given class appears in the ground truth (as opposed to either the ground truth or the prediction). This has the advantage that mistakes are only accounted for once, making the metric more optimistic than ours. On the other hand, information about the errors is lost since the error is only computed for the ground truth class independently of what the incorrectly predicted class is. 

On test datasets, the mDice (mean over the per-class dice scores \cite{milletari2016dice}) and Accuracy (overall percentage of correctly classified pixels, regardless of class) metrics are also reported.

\subsection{Experiment setting}

\textbf{\ac{sitta} training  set.} The \ac{sitta} training set for each model is derived from a set of 40 images from the segmentation model's training dataset extended with a set of 9 synthetic corruptions at three severity levels from \cite{hendrycks2019robustness} such as blur, noise or fog, simulating different domain shifts. The original images are also included since the \ac{tta} methods should not harm the model on source domain images. Details about the corruption can be found in Appendix \ref{sec:ap:corrs}. These synthetic datasets based on the GTA5 and COCO datasets are referred to as GTA5-C and COCO-C, respectively. Since the original images without corruption are also included, each \ac{sitta} training dataset consists of 1200 images (40 images, 9 + 1 corruption, three corruption levels). 

\textbf{\ac{sitta} hyper-parameters.} For each \ac{tta} method, optimizing all the network parameters or normalization parameters is only considered, resulting in at least two different setups for each method. Further, when applicable (the methods compute a segmentation loss based on masks, as opposed to another self-supervised loss such as the prediction entropy), the \ac{ce} and \ac{iou} losses are compared. While training segmentation models with a loss that takes class imbalance into account, such as the \ac{ce} and the Dice loss \cite{milletari2016dice}, is standard, \ac{tta} work on image segmentation has relied on cross-entropy, which is also suboptimal from the point of view of the evaluation metric. It is desirable to align the optimization metric with the evaluation metric as much as possible. This results in four setups for the \ac{ttaref}, \ac{ttapl}, and \ac{ttaaugco} methods. The learning rate and number of \ac{tta} iterations are considered from learning hyper-parameters. The maximum possible number of iterations is 10 to limit the computational requirements. Reasonable learning rate values are found via a grid search and then extended with other promising values based on the initial results.

\textbf{Shared implementation details.}
The refinement network architecture is a U-Net \cite{ronneberger2015unet} with an EfficentNet-B0 \cite{tan2019efficientnet} backbone pre-trained on ImageNet from the Timm library \cite{rw2019timm}.
It is trained with the AdamW \cite{Loshchilov2017DecoupledRegularization} optimizer with a learning rate of $1e^{-3}$ and the \acf{ce} loss. The SGD optimizer is used for the \ac{tta} since early experiments with AdamW showed a high divergence rate.

\subsection{Experiment results}
\begin{table*}[b]
\footnotesize
% \small
    \setlength{\tabcolsep}{2.4pt}
    \centering
\begin{tabular}{l|rr|rrrr|rrrr|rrrr|rr|rr}
\toprule
% \multicolumn{19}{c}{GTA5} \\ 
% \midrule
 \multicolumn{1}{c}{} & \multicolumn{2}{c}{Ent} & \multicolumn{4}{c}{PL} & \multicolumn{4}{c}{Ref} & \multicolumn{4}{c}{AugCo} & \multicolumn{2}{c}{Adv} & \multicolumn{2}{c}{dIoU} \\
\midrule
params & \multicolumn{1}{|c}{full} & \multicolumn{1}{c|}{norm} & \multicolumn{1}{c}{full} & \multicolumn{1}{c}{full} & \multicolumn{1}{c}{norm} & \multicolumn{1}{c|}{norm} & \multicolumn{1}{c}{full} & \multicolumn{1}{c}{full} & \multicolumn{1}{c}{norm} & \multicolumn{1}{c|}{norm} & \multicolumn{1}{c}{full} & \multicolumn{1}{c}{full} & \multicolumn{1}{c}{norm} & \multicolumn{1}{c|}{norm} & \multicolumn{1}{c}{full} & \multicolumn{1}{c|}{norm} & \multicolumn{1}{c}{full} & \multicolumn{1}{c}{norm} \\
loss & \multicolumn{1}{|c}{ent} & \multicolumn{1}{c|}{ent} & \multicolumn{1}{c}{ce} & \multicolumn{1}{c}{iou} & \multicolumn{1}{c}{ce} & \multicolumn{1}{c|}{iou} & \multicolumn{1}{c}{ce} & \multicolumn{1}{c}{iou} & \multicolumn{1}{c}{ce} & \multicolumn{1}{c|}{iou} & \multicolumn{1}{c}{ce} & \multicolumn{1}{c}{iou} & \multicolumn{1}{c}{ce} & \multicolumn{1}{c|}{iou} & \multicolumn{1}{c}{kl} & \multicolumn{1}{c|}{kl} & \multicolumn{1}{c}{-} & \multicolumn{1}{c}{-} \\
\midrule
NA & 35.18 & 35.18 & 35.18 & 35.18 & 35.18 & 35.18 & 35.18 & 35.18 & 35.18 & 35.18 & 35.18 & 35.18 & 35.18 & 35.18 & 35.20 & 35.20 & 35.18 & 35.18 \\
$\text{TTA}_{\alpha^*}$ & 35.18 & \underline{35.58} & 35.54 & \underline{37.21} & 35.60 & 37.09 & 35.18 & \textbf{38.69} & 36.88 & 36.50 & 35.27 & \underline{35.66} & 35.35 & 35.39 & 35.20 & 35.20 & 35.18 & 35.18 \\
$\Delta_{\text{ABS}}$ & $-\epsilon$ & 0.39 & 0.36 & 2.03 & 0.42 & 1.90 & $-\epsilon$ & 3.51 & 1.70 & 1.32 & 0.09 & 0.48 & 0.17 & 0.21 & $-\epsilon$ & $-\epsilon$ & $-\epsilon$ & $-\epsilon$ \\
% $\Delta_{\text{CORR}}$ & $-\epsilon$ & 1.76 & 1.62 & 9.04 & 1.85 & 8.47 & $-\epsilon$ & 15.61 & 7.56 & 5.89 & 0.40 & 2.14 & 0.74 & 0.95 & $-\epsilon$ & $-\epsilon$ & $-\epsilon$ & $-\epsilon$ \\
% $\Delta_{\text{TOTAL}}$ & $-\epsilon$ & 0.61 & 0.56 & 3.13 & 0.64 & 2.94 & $-\epsilon$ & 5.41 & 2.62 & 2.04 & 0.14 & 0.74 & 0.26 & 0.33 & $-\epsilon$ & $-\epsilon$ & $-\epsilon$ & $-\epsilon$ \\
\bottomrule
\end{tabular}

%  maybe we can exclude IoU??

\caption{$\text{m}\overline{\text{IoU}}_i$ results aggregated across corruptions and levels in the GTA5-C dataset, compared to non-adapted (NA) performance. The \ac{tta} hyper-parameters $\alpha^*$ were selected for overall best performance of each method. The \textbf{overall} and \underline{per-method} best results are highlighted. No positive hyper-parameters are denoted by $-\epsilon$ (the performance converges to 0 from below).
}
\label{tab:sem_seg:gta5c:valid_agg}
\end{table*}

\begin{figure*}[tb]
    \centering
        \includegraphics[keepaspectratio, width=\linewidth]{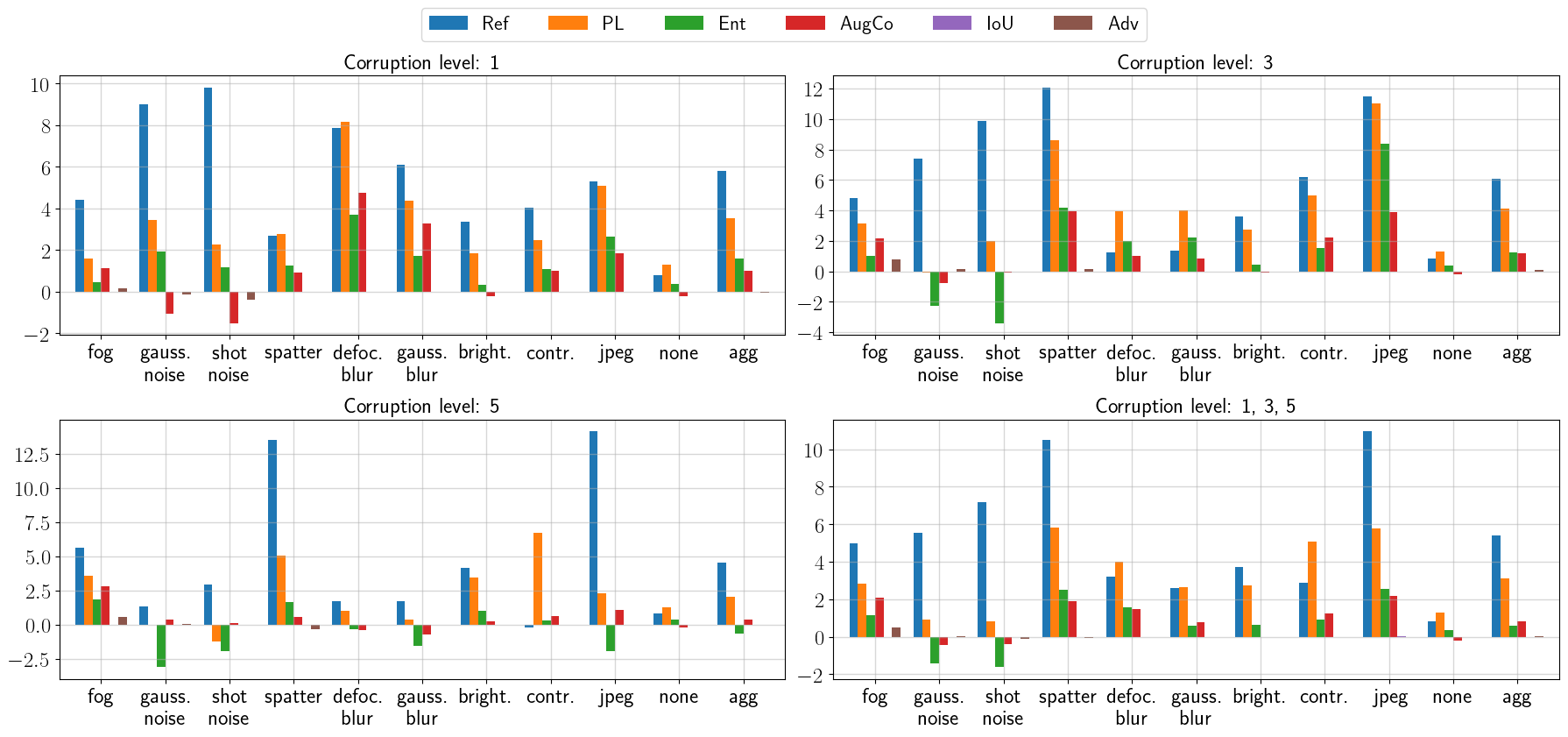}
        
      \caption{GTA5-C $\text{m}\overline{\text{IoU}}_i$  error reduction (\%) depending on corruption levels. \ac{tta} with overall optimal hyper-parameters for GTA5-C.}  
    \label{fig:gta5:sev_barplot_hparams_overall}
\end{figure*}

\begin{figure}[bth]
    \centering

    \begin{subfigure}[t]{0.49\textwidth}
      \includegraphics[keepaspectratio, width=\linewidth]{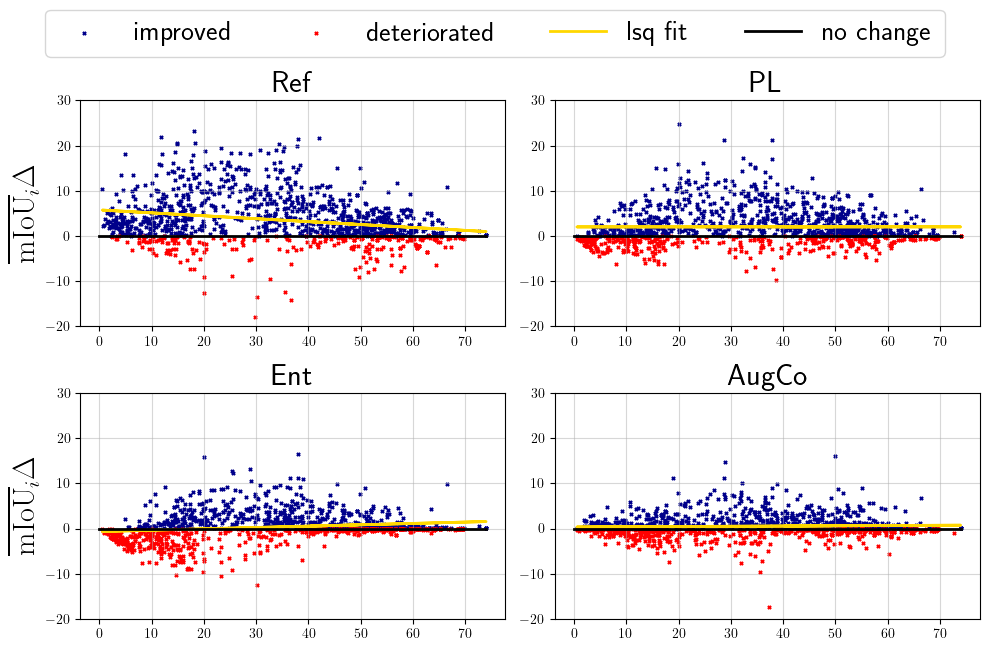}
      \caption{$\text{m}\overline{\text{IoU}}_i$ NA}
      \label{subfig:gta5:miou_na_tta}
    \end{subfigure}
    \hfill
    \begin{subfigure}[t]{0.49\textwidth}
      \includegraphics[keepaspectratio, width=\linewidth]{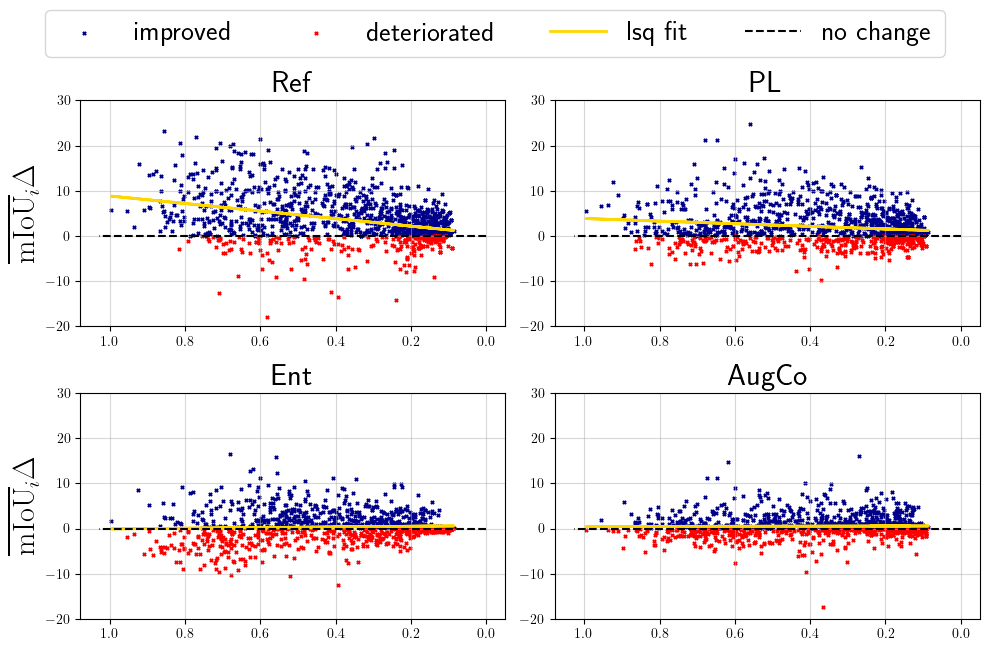}
      \caption{entropy NA}
        \label{subfig:gta5:ent_na_tta}
    \end{subfigure}

    \caption{The relationship between per-image scores (a) or entropy (b) before and the score after adaptation on the GTA5-C dataset. The difference between non-adapted (NA) $\text{m}\overline{\text{IoU}}_i$ or entropy and the $\text{m}\overline{\text{IoU}}_i$ after \ac{tta} is shown 
    ($\text{m}\overline{\text{IoU}}_i \Delta$). A least-squares line fitted to the points is shown in yellow.}
    \label{fig:gta5:na_tta}
\end{figure}

% \begin{figure}[bth]
%     \centering
%         \includegraphics[keepaspectratio, width=0.98\linewidth]{plots/gta5/per_image_miou_comp_deltas_na.png}
%     \caption{The relationship between per-image scores before and after adaptation on the GTA5-C dataset. The difference between non-adapted (NA) $\text{m}\overline{\text{IoU}}_i$ and the $\text{m}\overline{\text{IoU}}_i$ after \ac{tta} is shown 
%     ($\text{m}\overline{\text{IoU}}_i \Delta$). A least-squares line fitted to the points in yellow.}
%     \label{fig:gta5:miou_na_tta}
% \end{figure}

% \begin{figure}[bth]
%     \centering
%         \includegraphics[keepaspectratio, width=0.93\linewidth]{plots/gta5/per_image_miou_ent_delta_comp.png}
%     \caption{The relationship between per-image non-adapted (NA) prediction entropy and \ac{tta} $\text{m}\overline{\text{IoU}}_i$ improvement on the GTA5-C dataset. The difference between NA $\text{m}\overline{\text{IoU}}_i$ and the \ac{tta} $\text{m}\overline{\text{IoU}}_i$ is shown ($\text{m}\overline{\text{IoU}}_i \Delta$). Least-squares line fitted to the points in yellow.}
%     \label{fig:gta5:miou_ent_tta}
% \end{figure}
% \input{figures/gta5/tta_miou_tta_ent}
\textbf{GTA5 $\rightarrow$ Cityscapes, ACDC.}
This experiment explores the performance of the \ac{tta} methods on a model trained on a synthetic driving dataset, GTA5, evaluating on real-world driving datasets under different weather conditions.
The GTA5-pretrained model is the best-performing model of \cite{volpi2022road} (DeepLabV2).

Since current methods do not consider different hyper-parameters for individual images, a single set with overall best performance across all corruptions and corruption levels is selected. The aggregated results with these overall optimal hyper-parameters on the \ac{sitta} training set can be found in Table \ref{tab:sem_seg:gta5c:valid_agg}.
It can be observed that the biggest improvements are achieved either by \ac{ttapl} with \ac{iou} loss, optimizing normalization parameters only, or by \ac{ttaref} with \ac{iou} loss, optimizing all the parameters. The best-performing method is \ac{ttaref}, improving by 3.51 \% over the non-adapted baseline (NA). Other methods only marginally improve over NA or show no improvement at all. Optimizing \ac{ce} generally yields worse results than optimizing the \ac{iou}. While updating normalization parameters only may stabilize \ac{ttaent}, optimizing all the parameters is essential for optimal performance of \ac{ttaref}. For other methods, the difference is smaller - optimizing normalization parameters only is faster and thus recommended.

In Figure \ref{fig:gta5:sev_barplot_hparams_overall}, the total error reduction results with the same set of overall optimal hyper-parameters for each method are shown but for each corruption level and kind separately.
It can be seen that it is not possible to find a single set of hyper-parameters that would perform well across all the corruption levels with these methods. While all methods except for \ac{ttaaugco} improve performance on level 1 corruptions, from level 3, negative results can be observed more often and many methods deteriorate/only slightly improve on level 5. \ac{ttaref} outperforms the other methods on the majority of corruption kinds and corruption levels. The aggregated results across all corruptions showed that  the negative results are outweighed by the gains on level 1, resultin
g in overall positive results for most of the methods.

\begin{figure*}[tb]
    \centering
        \includegraphics[keepaspectratio, width=\linewidth]{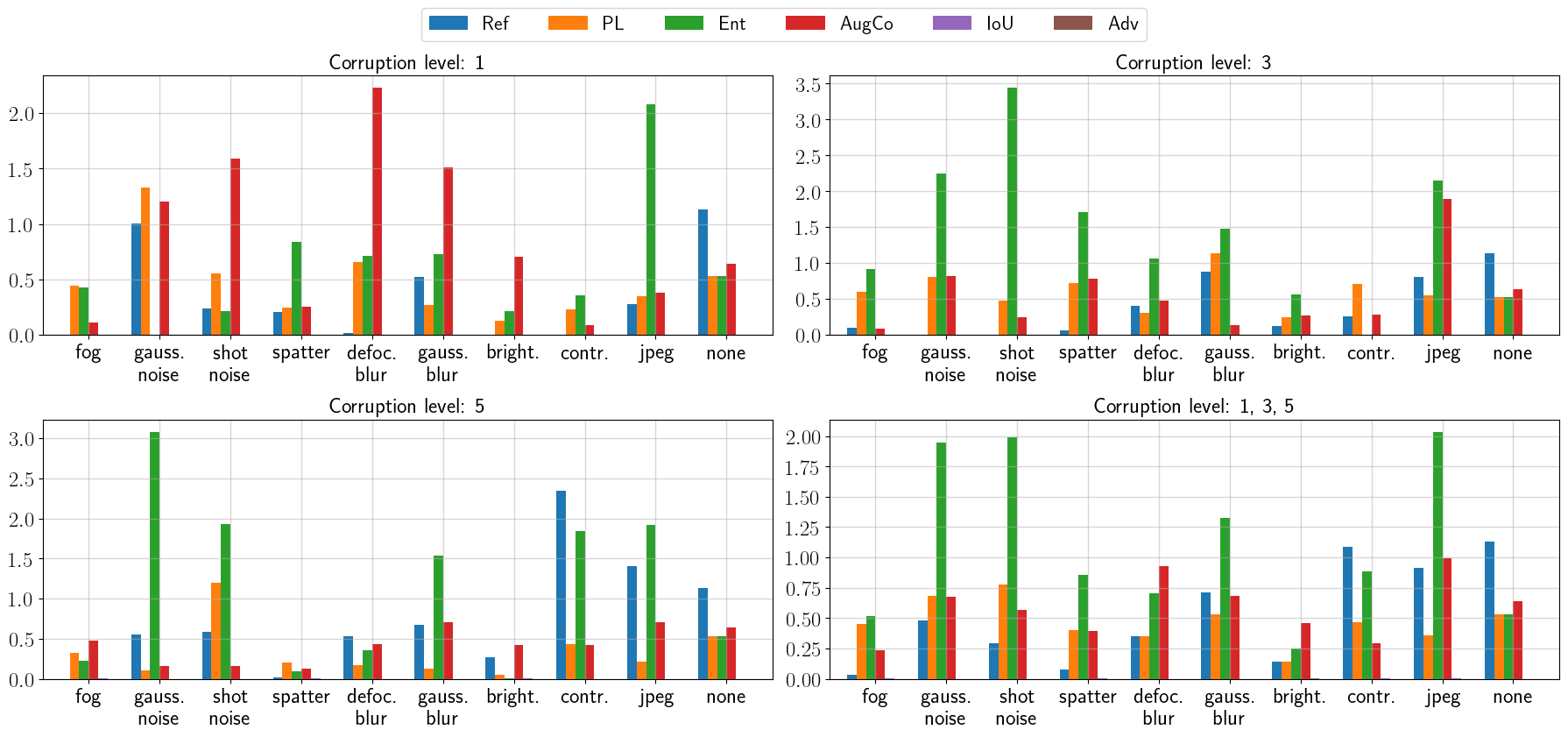}
    \caption{GTA5-C error reduction difference (\%) between overall optimal hyperparameters and hyper-parameters selected for each corruption kind separately. The hyper-parameters were selected on GTA5-C.}
    \label{fig:gta5:sev_barplot_hparams_each}
\end{figure*}

In Figure \ref{fig:gta5:sev_barplot_hparams_each}, it is shown that if one could select optimal hyper-parameters for each corruption kind and level, results would improve substantially for many of the corruptions. 
This analysis suggests that unless the domain shift is known in advance, strategies with method and hyper-parameter selection for each image separately should be explored.

Only the methods with overall positive TTA results are considered for further analysis, namely \ac{ttaref}, \ac{ttapl}, \ac{ttaaugco}, and \ac{ttaent}.
The relationship between the non-adapted (NA) performance and the performance improvement on individual images for different methods is visualized in Subfigure \ref{subfig:gta5:miou_na_tta}.
The analysis shows \ac{ttaref} outperforms other methods,
especially on images that had low initial $\text{m}\overline{\text{IoU}}_i$, while the performance of \ac{ttapl} is more consistent across all initial scores but not as powerful for initial low scores. While \ac{ttaent} makes performance worse for low initial scores and improves more as the initial score increases, \ac{ttaaugco} shows consistent improvements across all initial scores similarly to \ac{ttapl} but to a smaller extent.

\begin{figure}[bth]
    \centering
\begin{subfigure}[t]{0.483\textwidth}
      \includegraphics[keepaspectratio, width=\linewidth]{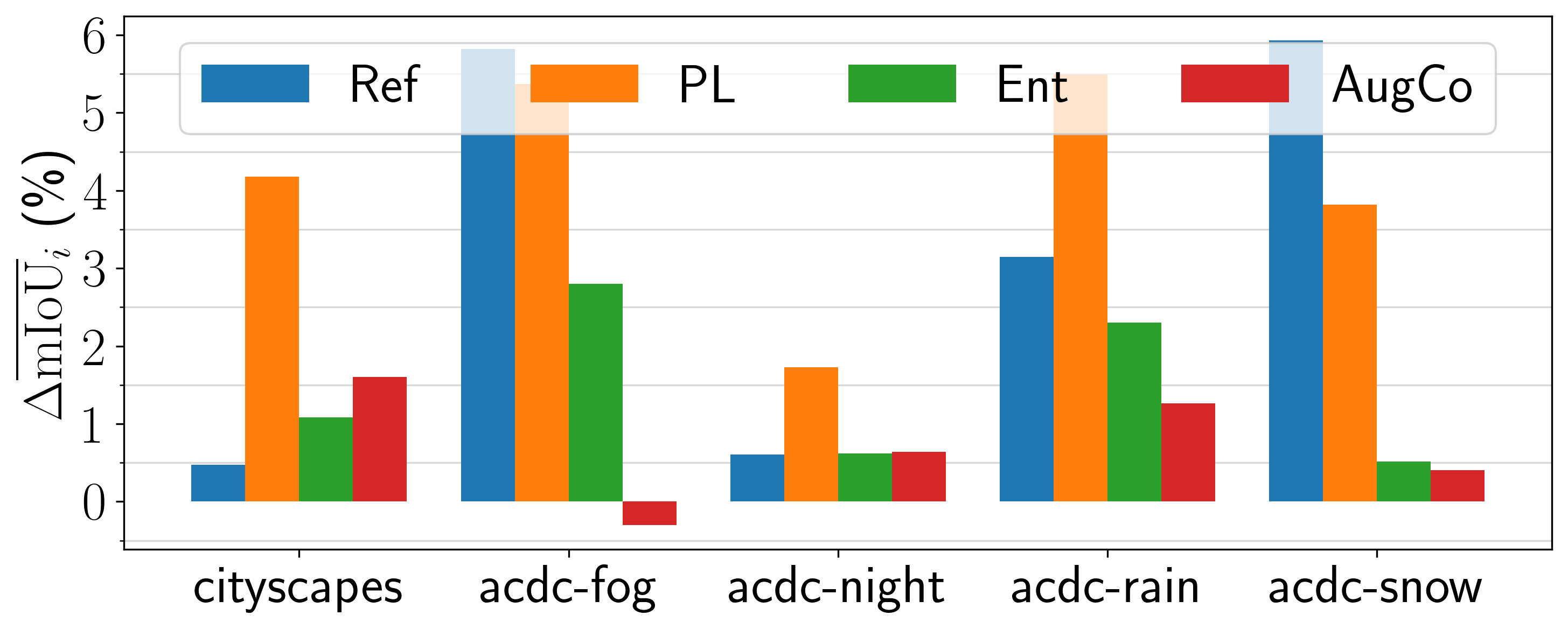}
      \caption{GTA5 model}
      \label{fig:driving:test_miou}
    \end{subfigure}
    \hfill
    \begin{subfigure}[t]{0.483\textwidth}
      \includegraphics[keepaspectratio, width=\linewidth]{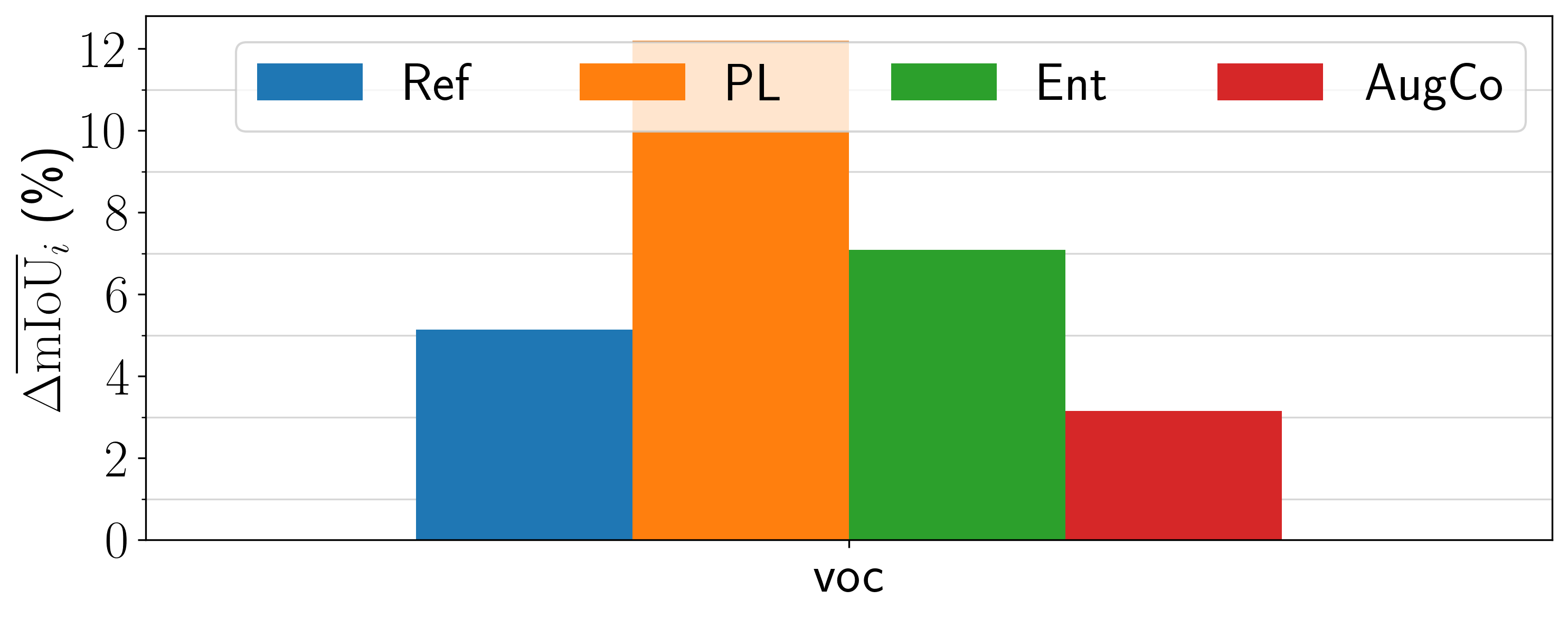}
      \caption{COCO model}
      \label{fig:coco:test_voc}
    \end{subfigure}
\caption{\ac{tta} $\text{m}\overline{\text{IoU}}_i$ error reduction (\%) on test datasets with hyper-parameters select on the \ac{tta} training datasets.
    }
\end{figure}

If the  $\text{m}\overline{\text{IoU}}_i$ for each image were known (the ground truth is necessary for its computation), it could be used to either select a method performing best on those values or to select hyper-parameters. In Subfigure \ref{subfig:gta5:ent_na_tta}, an analogous analysis is performed, replacing the  $\text{m}\overline{\text{IoU}}_i$ with segmentation prediction entropy, which does not require any supervision.
Similar results as with the $\text{m}\overline{\text{IoU}}_i$  can be observed. 

After selecting the best hyper-parameters for each method on the \ac{sitta} training set, the methods are evaluated on 5 test datasets: ACDC-Rain, ACDC-Fog, ACDC-Night, ACDC-Snow, and Cityscapes. The Cityscapes represent a domain shift from synthetic to real images, while ACDC datasets add adverse weather conditions, making the domain shift even greater. The first four datasets are created by splitting the ACDC dataset by different conditions. Each of the test sets consists of 500 images.
The test results are reported in Figure \ref{fig:driving:test_miou} and Table \ref{tab:sem_seg:driving:test_res}.
Similarly to \ac{sitta} training datasets, either \ac{ttaref} or \ac{ttapl} methods perform best, depending on the dataset. While not outperforming \ac{ttaref} on all the datasets, the performance of \ac{ttapl} is consistently better than the other methods while \ac{ttaref} is outperformed or matched by the other methods on Cityscapes and ACDC-night. Table \ref{tab:sem_seg:driving:test_res} shows that in standard metrics that do not consider per-image difficulty, \ac{ttaaugco} outperforms the other methods in many cases. 

\begin{table}[htb]
% \footnotesize
\small
    \centering
\renewcommand{\arraystretch}{1.15}

\begin{tabular}{llrrrrr}
\toprule
& & \multicolumn{5}{c}{method} \\
\cmidrule(lr){3-7} 
dataset & metric  &  \multicolumn{1}{c}{NA} &   \multicolumn{1}{c}{Ref} &    \multicolumn{1}{c}{PL} &  \multicolumn{1}{c}{Ent} &  \multicolumn{1}{c}{AugCo} \\
 % \cmidrule(lr){3-7} 
 \midrule
         
\multirow[c]{5}{*}{cityscapes} & $m\overline{\mathrm{IoU}}_i$ & 34.40 & 34.72 & \bfseries 37.14 & 35.12 & 35.52 \\
 & $m\overline{\mathrm{IoU}}_c$ & 28.71 & 28.65 & \bfseries 30.70 & 29.09 & 29.53 \\
 & mIoU & 42.90 & 40.37 & 43.28 & 43.41 & \bfseries 44.05 \\
 & mDice & 55.78 & 52.53 & 56.04 & 56.17 & \bfseries 56.89 \\
 & Accuracy & 86.54 & 83.92 & 87.15 & 87.30 & \bfseries 87.50 \\
\cline{1-7}
\multirow[c]{5}{*}{acdc-fog} & $m\overline{\mathrm{IoU}}_i$ & 32.03 & \bfseries 35.99 & 35.68 & 33.92 & 31.82 \\
 & $m\overline{\mathrm{IoU}}_c$ & 24.87 & 27.29 & \bfseries 27.52 & 26.00 & 24.69 \\
 & mIoU & 37.65 & 38.36 & \bfseries 39.39 & 39.15 & 37.61 \\
 & mDice & 51.42 & 51.12 & \bfseries 53.29 & 52.68 & 51.51 \\
 & Accuracy & 73.23 & \bfseries 84.11 & 75.51 & 75.91 & 66.63 \\
\cline{1-7}
\multirow[c]{5}{*}{acdc-night} & $m\overline{\mathrm{IoU}}_i$ & 13.60 & 14.12 & \bfseries 15.09 & 14.13 & 14.15 \\
 & $m\overline{\mathrm{IoU}}_c$ & 10.77 & 10.96 & \bfseries 11.53 & 10.68 & 11.01 \\
 & mIoU & 15.79 & 13.77 & 15.11 & 15.53 & \bfseries 16.25 \\
 & mDice & 24.38 & 21.29 & 23.49 & 23.74 & \bfseries 24.98 \\
 & Accuracy & 52.09 & 52.47 & 51.86 & 52.82 & \bfseries 53.09 \\
\cline{1-7}
\multirow[c]{5}{*}{acdc-rain} & $m\overline{\mathrm{IoU}}_i$ & 33.52 & 35.60 & \bfseries 37.16 & 35.05 & 34.50 \\
 & $m\overline{\mathrm{IoU}}_c$ & 26.15 & 27.40 & \bfseries 28.47 & 26.89 & 26.73 \\
 & mIoU & 36.93 & 36.21 & 37.61 & 37.53 & \bfseries 37.98 \\
 & mDice & 48.92 & 47.52 & 49.44 & 49.23 & \bfseries 50.25 \\
 & Accuracy & 84.56 & 84.22 & 85.65 & \bfseries 86.13 & 84.74 \\
\cline{1-7}
\multirow[c]{5}{*}{acdc-snow} & $m\overline{\mathrm{IoU}}_i$ & 31.54 & \bfseries 35.60 & 34.15 & 31.87 & 31.81 \\
 & $m\overline{\mathrm{IoU}}_c$ & 25.28 & \bfseries 28.09 & 27.16 & 25.38 & 25.45 \\
 & mIoU & 35.30 & \bfseries 37.89 & 36.64 & 35.34 & 35.40 \\
 & mDice & 48.46 & \bfseries 50.35 & 50.00 & 48.51 & 48.53 \\
 & Accuracy & 73.17 & \bfseries 81.52 & 74.48 & 73.66 & 73.45 \\
\bottomrule

        \end{tabular}
        
        \caption{ACDC and Cityscapes test datasets results. Hyper-parameters were selected for overall best performance on GTA5-C. Best results for each dataset and metric are \textbf{highlighted}.
        }
        \label{tab:sem_seg:driving:test_res}
\end{table}

The inconsistencies of results between \ac{sitta} training and test suggest that unless the domain shift conditions are known in advance, it is difficult to select hyper-parameters based on a general \ac{sitta} training set. 

\textbf{COCO $\rightarrow$ VOC}.
In this experiment, the performance of \ac{tta} methods is studied on a model trained on the COCO dataset and evaluated on the VOC dataset.
The segmentation model is an official Torchvision DeepLabV3 model with a Resnet50 backbone trained on the COCO dataset with a subset of 20 VOC classes. In contrast to previous experiments, it is a real-to-real dataset domain shift.
The results of different methods with parameters selected for the overall best performance across all corruptions and levels can be found in Table
\ref{tab:sem_seg:cococ:valid_agg}.
Biggest improvements are obtained by the \ac{ttapl} and \ac{ttaref} methods.  \ac{ttapl} outperforms \ac{ttaref}, in contrast to the GTA5-C experiments. The best improvement is by 3.28 \%, reducing the total segmentation error by 7.3 \%.  Consistently with previous experiments, best results are achieved with the \ac{iou} loss, outperforming \ac{ce} in all cases. In contrast to GTA5-C, \ac{ttaent} achieves better results when optimizing all the network parameters, as opposed to optimizing only the normalization layer parameters. The same holds for \ac{ttapl}. For \ac{ttaref}, optimizing all the parameters is again important. Other methods' improvements over the non-adapted baseline are marginal.

\begin{table*}[h]
\footnotesize
% \small
    \setlength{\tabcolsep}{2.4pt}
    \centering
\begin{tabular}{l|rr|rrrr|rrrr|rrrr|rr|rr}
\toprule
% \multicolumn{19}{c}{GTA5} \\ 
% \midrule
 \multicolumn{1}{c}{} & \multicolumn{2}{c}{Ent} & \multicolumn{4}{c}{PL} & \multicolumn{4}{c}{Ref} & \multicolumn{4}{c}{AugCo} & \multicolumn{2}{c}{Adv} & \multicolumn{2}{c}{dIoU} \\
\midrule
params & \multicolumn{1}{|c}{full} & \multicolumn{1}{c|}{norm} & \multicolumn{1}{c}{full} & \multicolumn{1}{c}{full} & \multicolumn{1}{c}{norm} & \multicolumn{1}{c|}{norm} & \multicolumn{1}{c}{full} & \multicolumn{1}{c}{full} & \multicolumn{1}{c}{norm} & \multicolumn{1}{c|}{norm} & \multicolumn{1}{c}{full} & \multicolumn{1}{c}{full} & \multicolumn{1}{c}{norm} & \multicolumn{1}{c|}{norm} & \multicolumn{1}{c}{full} & \multicolumn{1}{c|}{norm} & \multicolumn{1}{c}{full} & \multicolumn{1}{c}{norm} \\
loss & \multicolumn{1}{|c}{ent} & \multicolumn{1}{c|}{ent} & \multicolumn{1}{c}{ce} & \multicolumn{1}{c}{iou} & \multicolumn{1}{c}{ce} & \multicolumn{1}{c|}{iou} & \multicolumn{1}{c}{ce} & \multicolumn{1}{c}{iou} & \multicolumn{1}{c}{ce} & \multicolumn{1}{c|}{iou} & \multicolumn{1}{c}{ce} & \multicolumn{1}{c}{iou} & \multicolumn{1}{c}{ce} & \multicolumn{1}{c|}{iou} & \multicolumn{1}{c}{kl} & \multicolumn{1}{c|}{kl} & \multicolumn{1}{c}{-} & \multicolumn{1}{c}{-} \\
\midrule
NA & 55.01 & 55.01 & 55.01 & 55.01 & 55.01 & 55.01 & 55.01 & 55.01 & 55.01 & 55.01 & 55.01 & 55.01 & 55.01 & 55.01 & 55.16 & 55.16 & 55.01 & 55.01 \\
$\text{TTA}_{\theta^*}$ & \underline{56.97} & 56.75 & 57.17 & 57.99 & 57.10 & \textbf{58.30} & 56.24 & \underline{57.31} & 56.56 & 57.16 & 55.40 & 55.59 & 55.30 & \underline{56.30} & 55.16 & 55.16 & 55.61 & \underline{55.74} \\
$\Delta_{\text{ABS}}$ & 1.96 & 1.74 & 2.16 & 2.98 & 2.09 & 3.28 & 1.23 & 2.30 & 1.55 & 2.15 & 0.39 & 0.58 & 0.29 & 1.29 & $-\epsilon$ & $-\epsilon$ & 0.60 & 0.73 \\
% $\Delta_{\text{CORR}}$ & 16.98 & 15.08 & 18.70 & 25.79 & 18.14 & 28.47 & 10.68 & 19.92 & 13.40 & 18.61 & 3.38 & 5.03 & 2.48 & 11.20 & $-\epsilon$ & $-\epsilon$ & 5.16 & 6.32 \\
% $\Delta_{\text{TOTAL}}$ & 4.35 & 3.87 & 4.80 & 6.61 & 4.65 & 7.30 & 2.74 & 5.11 & 3.44 & 4.77 & 0.87 & 1.29 & 0.63 & 2.87 & $-\epsilon$ & $-\epsilon$ & 1.32 & 1.62 \\
\bottomrule
\end{tabular}

%  maybe we can exclude IoU??

\caption{$\text{m}\overline{\text{IoU}}_i$ results aggregated across corruptions and levels in the COCO-C dataset, compared to non-adapted (NA) performance. The \ac{tta} hyper-parameters $\alpha^*$ were selected for overall best performance of each method. The \textbf{overall} and \underline{per-method} best results are highlighted. No positive hyper-parameters are denoted by $-\epsilon$ (the performance converges to 0 from below).}
\label{tab:sem_seg:cococ:valid_agg}
\end{table*}

\begin{figure*}[bt]
    \centering
        \includegraphics[keepaspectratio, width=\linewidth]{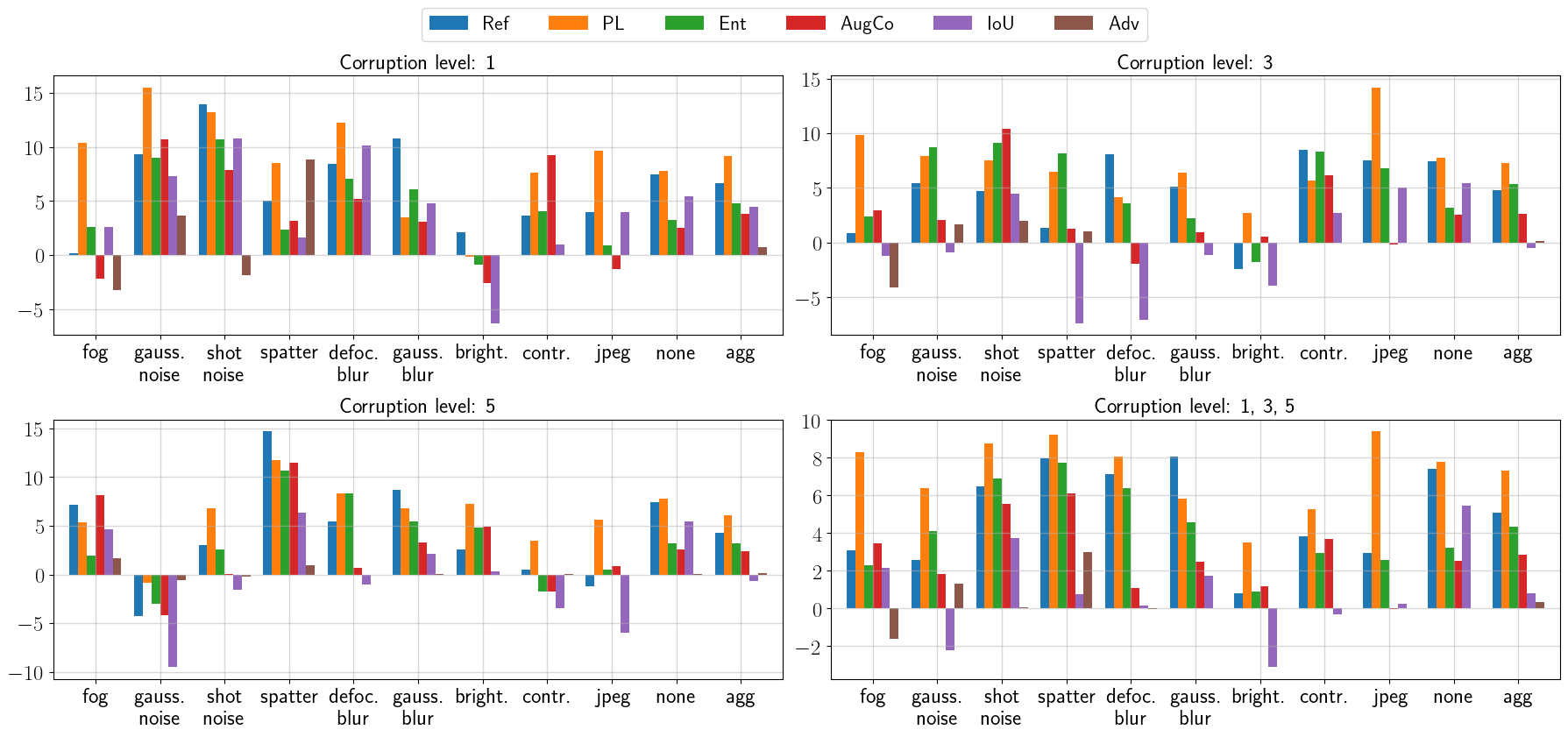}
\caption{COCO-C $\text{m}\overline{\text{IoU}}_i$  error reduction (\%) depending on corruption levels. \ac{tta} with overall optimal hyper-parameters for COCO-C.}  
    \label{fig:coco:sev_barplot_hparams_overall}
\end{figure*}

\begin{figure}[bth]
    \centering

    \begin{subfigure}[t]{0.49\textwidth}
      \includegraphics[keepaspectratio, width=\linewidth]{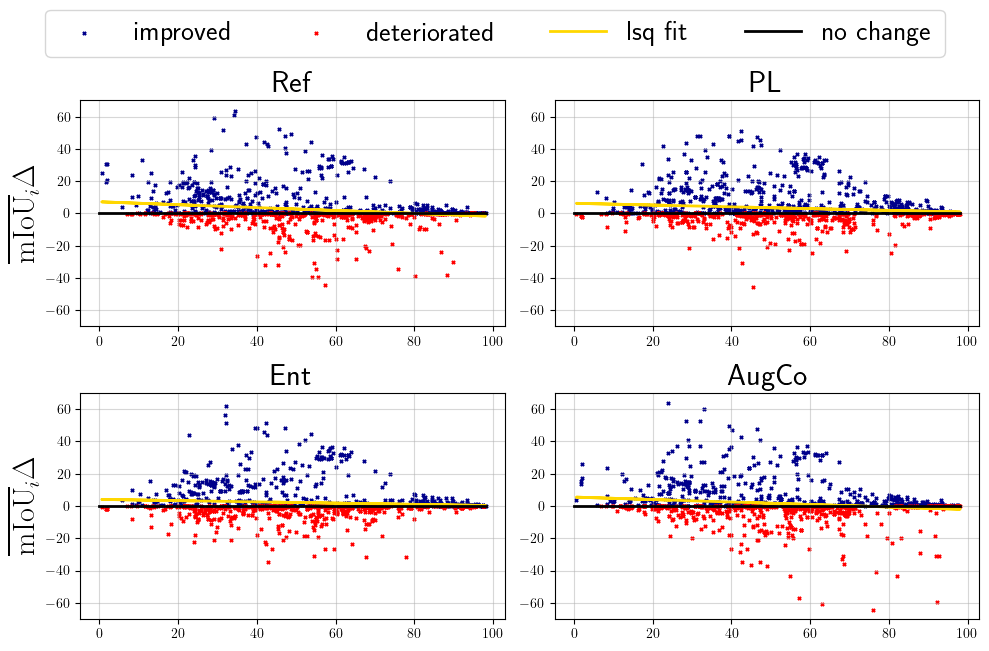}
      \caption{$\text{m}\overline{\text{IoU}}_i$ NA}
      \label{subfig:cococ:miou_na_tta}
    \end{subfigure}
    \hfill
    \begin{subfigure}[t]{0.49\textwidth}
      \includegraphics[keepaspectratio, width=\linewidth]{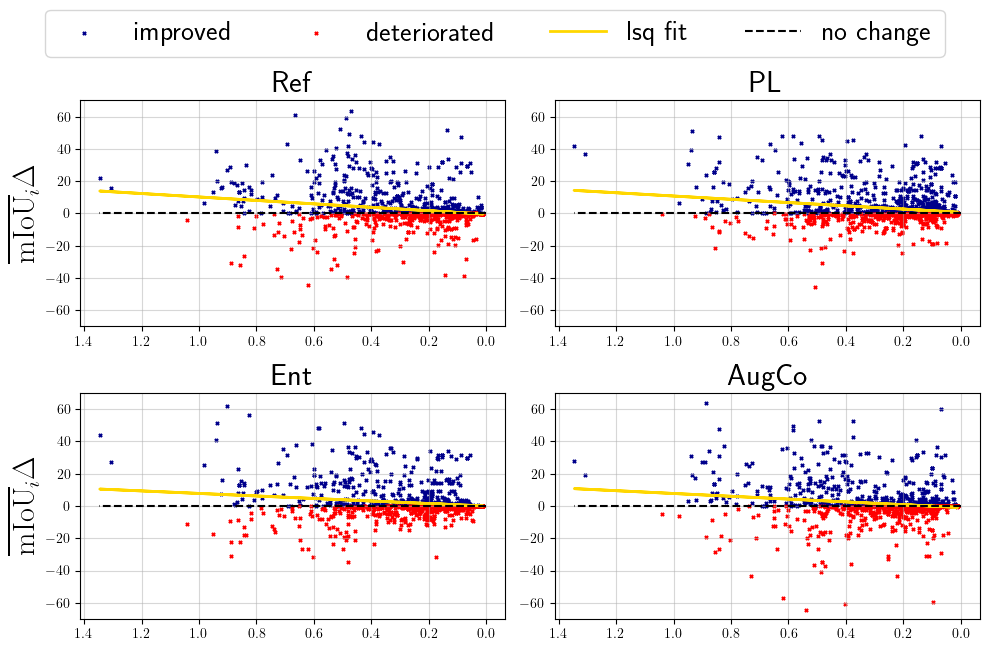}
      \caption{entropy NA}
        \label{subfig:cococ:ent_na_tta}
    \end{subfigure}

    \caption{The relationship between per-image scores (a) or entropy (b) before and the score after adaptation on the COCO-C dataset. The difference between non-adapted (NA) $\text{m}\overline{\text{IoU}}_i$ or entropy and the $\text{m}\overline{\text{IoU}}_i$ after \ac{tta} is shown 
    ($\text{m}\overline{\text{IoU}}_i \Delta$). A least-squares line fitted to the points is shown in yellow.}
    \label{fig:cococ:na_tta}
\end{figure}

% \begin{figure}[bth]
%     \centering
%         \includegraphics[keepaspectratio, width=0.98\linewidth]{plots/coco/per_image_miou_comp_deltas_na.png}
%     \caption{The relationship between per-image scores before and after adaptation on the COCO-C dataset. The difference between non-adapted (NA) $\text{m}\overline{\text{IoU}}_i$ and the $\text{m}\overline{\text{IoU}}_i$ after \ac{tta} is shown 
%     ($\text{m}\overline{\text{IoU}}_i \Delta$). A least-squares line fitted to the points in yellow.}
%     \label{fig:coco:miou_na_tta}
% \end{figure}

% \begin{figure}[bth]
%     \centering
%         \includegraphics[keepaspectratio, width=0.93\linewidth]{plots/coco/per_image_miou_ent_delta_comp.png}
%     \caption{The relationship between per-image non-adapted (NA) prediction entropy and \ac{tta} $\text{m}\overline{\text{IoU}}_i$ improvement on the COCO-C dataset. The difference between NA $\text{m}\overline{\text{IoU}}_i$ and the \ac{tta} $\text{m}\overline{\text{IoU}}_i$ is shown ($\text{m}\overline{\text{IoU}}_i \Delta$). Least-squares line fitted to the points in yellow.}
%     \label{fig:coco:miou_ent_tta}
% \end{figure}
% \input{figures/coco/tta_miou_tta_ent}

The total error reduction results with a single set of optimal hyper-parameters for each method are reported for each corruption level and kind in Figure \ref{fig:coco:sev_barplot_hparams_overall}. The results slightly differ from those for the GTA5-C, as in this case, \ac{ttapl} is consistently the best method, only rarely outperformed by \ac{ttaref} or \ac{ttaaugco}. Negative results on some corruption kinds are already reported for level 1 corruptions.

\begin{figure*}[tb]
    \centering
        \includegraphics[keepaspectratio, width=\linewidth]{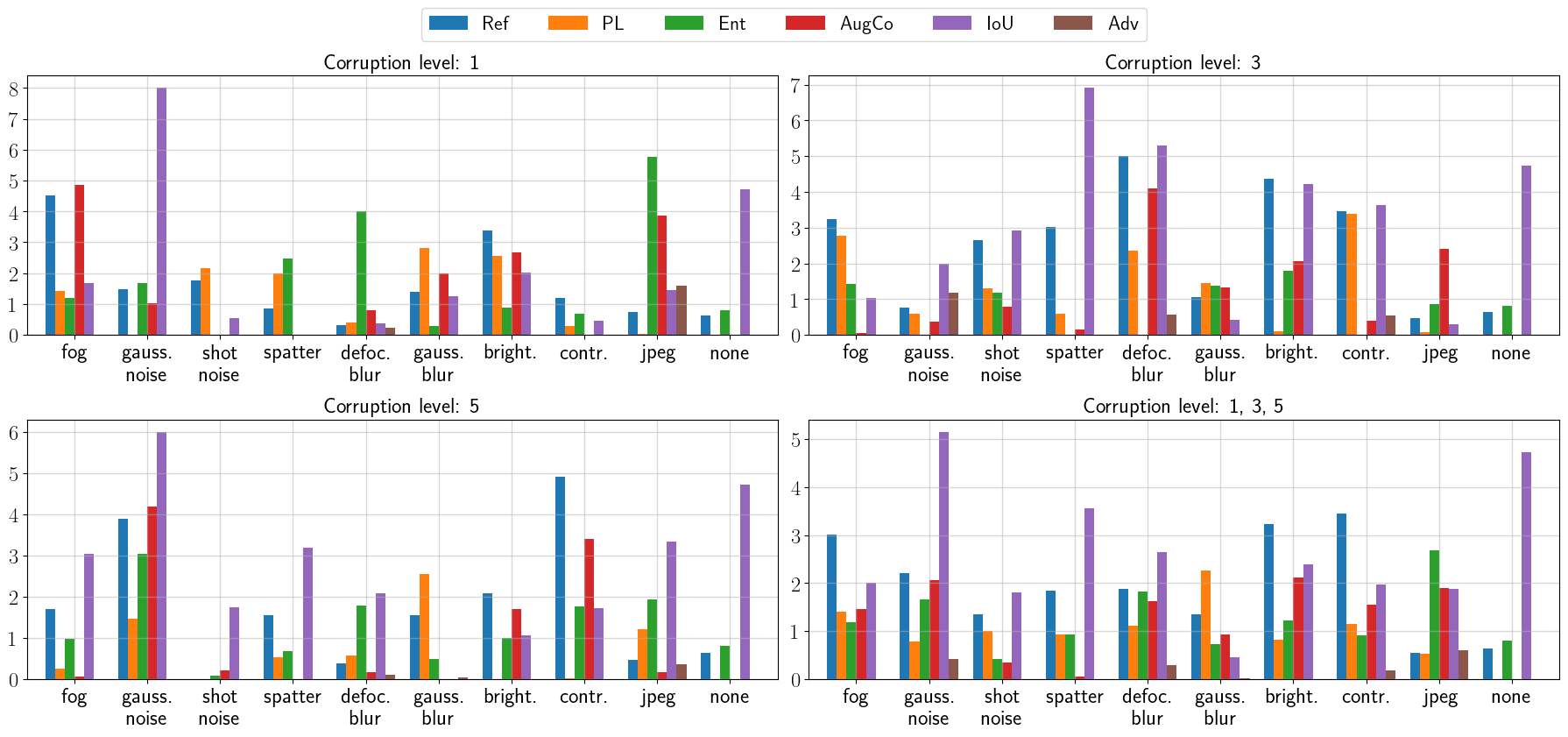}
    \caption{COCO-C error reduction difference (\%) between overall optimal hyperparameters and hyper-parameters selected for each corruption kind separately. The hyper-parameters were selected on the COCO-C.}
    \label{fig:coco:sev_barplot_hparams_each}
\end{figure*}

In Figure \ref{fig:coco:sev_barplot_hparams_each}, the results with optimal hyper-parameters for each method, corruption kind and level are shown. The results again improve substantially but this time, the differences between methods are smaller.  Interestingly, the \ac{ttaiou} method performs much stronger than in the GTA5-C experiments. 

\begin{figure}[bth]
    \centering

    \begin{subfigure}[t]{0.483\textwidth}
      \includegraphics[keepaspectratio, width=1\linewidth]{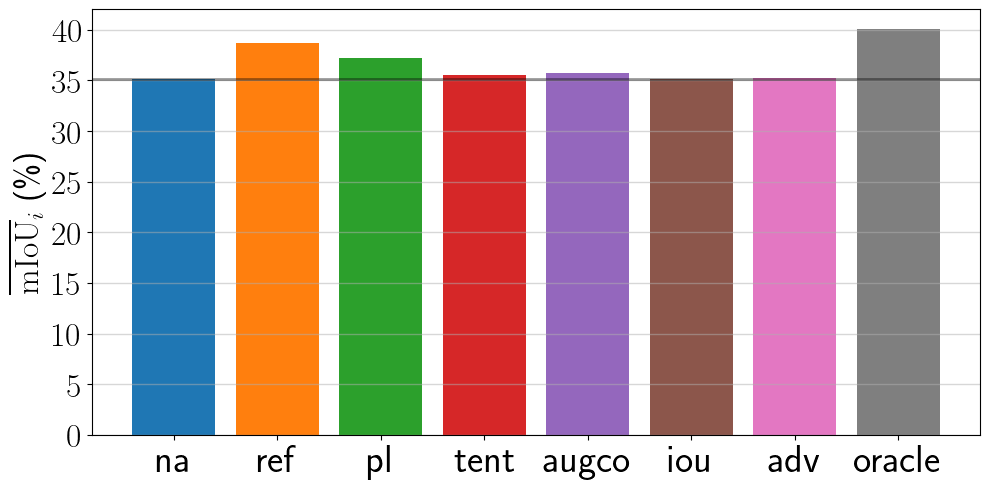}
    \caption{Comparison of the overall performance of different methods on the GTA5-C validation set.
    }
    \label{fig:gta:val_method_comp}
    \end{subfigure}
    \hfill
    \begin{subfigure}[t]{0.5\textwidth}
      \includegraphics[keepaspectratio, width=1\linewidth]{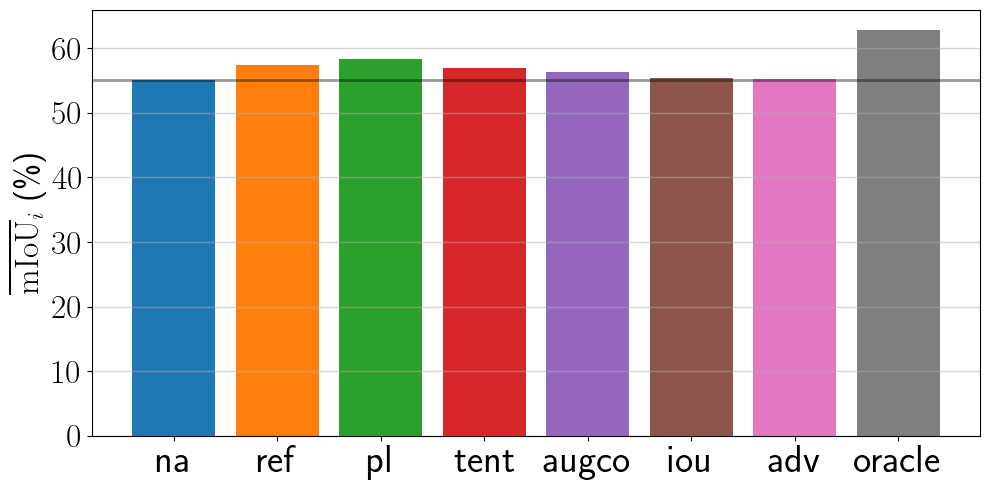}
    \caption{Comparison of the overall performance of different methods on the COCO-C validation set.
    }
    \label{fig:coco:val_method_comp}
    \end{subfigure}

    \caption{ The difference between non-adapted (NA) $\text{m}\overline{\text{IoU}}_i$ .}
    \label{fig:gta5:na_tta}
\end{figure}

Figure \ref{fig:coco:val_method_comp} compares the overall method performance on the \ac{sitta} training set. An oracle option is introduced where the method with the best results is picked for each image. There is a significant gap between the oracle and other methods, which further highlights that different methods are good in various cases and understanding the strengths of each methods can lead to significantly improved performance.

% (GTA5-C) and  \(COCO-C)
% Further, while on the GTA5-C dataset, \ac{ttaref} sinigificantly outperforms the other methods, \ac{ttapl} is better in the VOC-C benchmark.


Again, only \ac{ttaref}, \ac{ttapl}, \ac{ttaaugco}, and \ac{ttaent} are used for further analysis.
The relationship between the non-adapted (NA) performance and the performance improvement on individual images for different methods is visualized in Subfigure \ref{subfig:cococ:miou_na_tta}.
The distribution of initial non-adapted $\text{m}\overline{\text{IoU}}_i$ is different. The initial model is stronger than the GTA5 model. All methods show similar behavior - more improvement is achieved on images with a lower initial score.

The relationship of segmentation prediction entropy and $\text{m}\overline{\text{IoU}}_i$ improvement by adaptation is shown in Subfigure \ref{subfig:cococ:ent_na_tta}, supporting the notion that the entropy of prediction before adaptation is a good proxy for $\text{m}\overline{\text{IoU}}_i$.

\begin{table}[htb]
% \footnotesize
\small
    \setlength{\tabcolsep}{4pt}
    \centering

        \begin{tabular}{lrrrrr}
        \toprule
        & \multicolumn{5}{c}{method} \\
        \cmidrule(lr){2-6} 
         metric  &    \multicolumn{1}{c}{NA} &   \multicolumn{1}{c}{Ref} &    \multicolumn{1}{c}{PL} &  \multicolumn{1}{c}{Ent} &  \multicolumn{1}{c}{AugCo} \\
         % \cmidrule(lr){2-6} 
         \midrule
         
$m\overline{\mathrm{IoU}}_i$ & 77.00 & 77.62 & \bfseries 79.67 & 78.85 & 77.75 \\
$m\overline{\mathrm{IoU}}_c$ & 63.38 & 63.38 & \bfseries 66.90 & 65.38 & 63.10 \\
mIoU & 77.20 & \bfseries 77.78 & 77.55 & 77.01 & 76.01 \\
mDice & 86.16 & \bfseries 86.66 & 86.47 & 86.05 & 85.54 \\
Accuracy & 94.69 & \bfseries 95.18 & 94.68 & 94.69 & 94.76 \\
\bottomrule
        \end{tabular}
        
        \caption{VOC test dataset results. Hyper-parameters were selected for overall best performance on COCO-C. Best results for each dataset and metric are \textbf{highlighted.}
        }
        \label{tab:sem_seg:voc:test_res}
\end{table}

The results on the VOC test set are shown in Figure \ref{fig:coco:test_voc} and Table \ref{tab:sem_seg:voc:test_res}. \ac{ttapl} outperforms other methods in the proposed metrics, all the methods improved over the non-adapted baseline. When considering standard metrics which do not consider per-image difficulty, \ac{ttaref} is best. 

Additional results can be found in Appendix \ref{sec:ap:exps}.

\section{Conclusions and limitations}

% \item We conduct a comparative study of six \ac{tta} techniques run in \ac{sitta} mode for image segmentation: Two established baselines, two adapted state-of-the-art methods from image classification and continual \ac{tta}, and two proposed methods. 
    % \item Novel methods adapting ideas from unsupervised domain adaptation and medical imaging \ac{tta} to non-medical image segmentation are introduced, filling a gap in exploring diverse self-supervised loss functions. The method outperforms the other methods on multiple test datasets and is shown to be powerful on the images with the worst segmentation performance, as measured by \ac{iou}.
    % % A novel adversarial refinement module training for \acl{ttaref} \acl{tta}.
    % \item Improvements of baselines in the single-image setup by replacing \ac{ce} with the \ac{iou} loss.
    % The performance of pseudo-labelling is improved by 3.51 \% and 3.28 \% on GTA5-C and COCO-C validation sets while with \ac{ce} loss, the improvements are by 1.7 \%  and 2.16 \% only, respectively.
    % \item The first work shows the potential of \ac{sitta} for segmentation, an underexplored setup essential in applications with strict data governance standards or high variability among individual images.
    
This work investigated the performance of six \acf{sitta} methods on semantic segmentation problems. Two established baselines, two adapted state-of-the-art methods from image classification and continual \ac{tta}, and two novel methods were considered. We designed a framework that allows for hyper-parameter tuning and analysis of performance under diverse conditions on synthetic data inspired by \cite{hendrycks2019robustness} that can be derived from an arbitrary training dataset.  We evaluated the methods on real-world datasets.

Experiments on driving-scene datasets dominate the segmentation \ac{tta} literature. We followed this practice and experimented with models pretrained on the synthetic GTA5 dataset, evaluating on real-world driving scenes, including datasets with adverse weather conditions such as rain and fog. The common practice is to tune \ac{tta} hyperparameters on the Synthia \cite{ros2016synthia} datasets, however, this dataset contains the same weather and day-night conditions as the test datasets. Also, it is limited to driving datasets with the same set of classes. For this reason, we tuned hyper-parameters on a synthetic dataset derived from the training dataset by applying different corruptions, which gives us control over the domain shift conditions, facilitates analysis and is applicable to any existing dataset. We also added a novel benchmark where we evaluated a COCO-pretrained model on the VOC dataset, focusing on common objects.

We proposed two new methods inspired by ideas from unsupervised domain adaptation and medical imaging \ac{tta}. The stronger of the methods based on a learnt refinement module performed best on multiple of the test datasets and we showed that it is powerful on the images with the worst segmentation performance, as measured by \ac{iou}. We also showed that if we replace the \ac{iou} performance with the entropy of the predicted segmentation mask, which does not require the ground truth to be known, the same behaviour can be observed. This could be used to choose an appropriate method/hyper-parameters in future work.

We explored the effect of previously neglected design choices. Training with a loss function that accounts for class imbalance, a known issue of image segmentation datasets, such as the \ac{iou} or dice loss, is standard when training image segmentation models. When considering small batch sizes or even a single image, the class imbalance further increases, but \ac{sitta} methods implement baselines such as pseudo-labelling with the \ac{ce} loss or entropy minimization with equal weight for all pixels. 
Our results revealed that while \ac{sitta} in the standard setting with \ac{ce} loss did not improve performance much, substituting the \ac{ce} with \ac{iou} improves performance substantially. The performance of pseudo-labelling was improved by 3.51 \% and 3.28 \% on GTA5-C and COCO-C validation sets while with \ac{ce} loss, the improvements were by 1.7 \% 2.16 \% only, respectively. 
The experiments on whether to update all or normalization parameters only show this design choice significantly impacts results but the right option depends on the settings (method, dataset) - it is an important hyper-parameter to consider. Further, we find that entropy minimization, often reported as unstable for small batch sizes, performs well when the batch-normalization mean and variance are not updated at test time and only the affine parameters of the normalization layers are optimized.

In the GTA5-C synthetic datasets experiments, the refinement \ac{sitta} dominates, followed by the pseudo-labelling baseline.
While the refinement is significantly better on some of the real-world test datasets, on other ones, pseudo-labelling performs best.
In the COCO-C experiments, the top performers swap places: Pseudo-labelling is followed by refinement.
On the test dataset, pseudo-labelling remains the best.  While the other methods do not perform very well when \ac{tta} hyperparameters are optimized across many different corruption kinds, their performance improves when tuning them to specific kinds of domain shift.
While there is not a single method performing best over all the test datasets, our results highlight the potential of \ac{sitta} for semantic segmentation.

\textbf{Limitations.} To limit the scope of the study, we only focused on adaptation with self-supervised loss functions and no reliance on batch normalization layers. While these methods tend to perform the best, their iterative optimization comes at an increased computational time.
Methods alleviating this burden should be explored, such as only adapting to informative samples or methods inspired by efficient model finetuning.

The synthetic validation set created by applying artificial corruptions to the training set may not cover the complexity of real-world domain shifts. Label shift, common in real-world datasets, is not considered - the solutions are typically complementary. 

Finally, only two models were considered and the effect of different model architectures on the individual methods is not known. While our work improves the understanding of \ac{tta} for semantic segmentation methods, a benchmark for fair and thorough evaluation of the methods is still missing.

\subsection*{Acknowledgements}
We greatly appreciate the help of Tomas Jenicek and Ondrej Tybl with proofreading of the manuscript.

Klara Janouskova was supported by the SGS23/173/OHK3/3T/13 research grant, and Jiri Matas was supported by the Technology Agency of the Czech Republic, project No. SS73020004 and project FACIS No. VJ02010041. Tamir Shor and Chaim Baskin were supported by NOFAR MAGNET number 78732 by the Israel Innovation Authority.

% * so that it is not numbered and doesn't look like a part of the previous section

% \subsubsection*{Broader Impact Statement}
% In this optional section, TMLR encourages authors to discuss possible repercussions of their work,
% notably any potential negative impact that a user of this research should be aware of. 
% Authors should consult the TMLR Ethics Guidelines available on the TMLR website
% for guidance on how to approach this subject.

% Maybe we could discuss a bit the 'Test-Time Poisoning Attacks Against Test-Time Adaptation Models' work and similar.

\bibliography{references, main}

\begin{thebibliography}{53}
\providecommand{\natexlab}[1]{#1}
\providecommand{\url}[1]{\texttt{#1}}
\expandafter\ifx\csname urlstyle\endcsname\relax
  \providecommand{\doi}[1]{doi: #1}\else
  \providecommand{\doi}{doi: \begingroup \urlstyle{rm}\Url}\fi

\bibitem[Aquino et~al.(2017)Aquino, Gutoski, Hattori, and Lopes]{aquino2017effect}
N~Romero Aquino, Matheus Gutoski, Leandro~T Hattori, and Heitor~S Lopes.
\newblock The effect of data augmentation on the performance of convolutional neural networks.
\newblock \emph{Braz. Soc. Comput. Intell}, 2017.

\bibitem[Ba et~al.(2016)Ba, Kiros, and Hinton]{ba2016layer}
Jimmy~Lei Ba, Jamie~Ryan Kiros, and Geoffrey~E Hinton.
\newblock Layer normalization.
\newblock \emph{arXiv preprint arXiv:1607.06450}, 2016.

\bibitem[Bartler et~al.(2022)Bartler, B{\"u}hler, Wiewel, D{\"o}bler, and Yang]{bartler2022mt3}
Alexander Bartler, Andre B{\"u}hler, Felix Wiewel, Mario D{\"o}bler, and Bin Yang.
\newblock Mt3: Meta test-time training for self-supervised test-time adaption.
\newblock In \emph{International Conference on Artificial Intelligence and Statistics}, pp.\  3080--3090. PMLR, 2022.

\bibitem[Chen et~al.(2022)Chen, Wang, Darrell, and Ebrahimi]{chen2022contrastive}
Dian Chen, Dequan Wang, Trevor Darrell, and Sayna Ebrahimi.
\newblock Contrastive test-time adaptation.
\newblock In \emph{Proceedings of the IEEE/CVF Conference on Computer Vision and Pattern Recognition}, pp.\  295--305, 2022.

\bibitem[Croce et~al.(2020)Croce, Andriushchenko, Sehwag, Debenedetti, Flammarion, Chiang, Mittal, and Hein]{croce2020robustbench}
Francesco Croce, Maksym Andriushchenko, Vikash Sehwag, Edoardo Debenedetti, Nicolas Flammarion, Mung Chiang, Prateek Mittal, and Matthias Hein.
\newblock Robustbench: a standardized adversarial robustness benchmark.
\newblock \emph{arXiv preprint arXiv:2010.09670}, 2020.

\bibitem[Fournier et~al.(1982)Fournier, Fussell, and Carpenter]{fournier1982computer}
Alain Fournier, Don Fussell, and Loren Carpenter.
\newblock Computer rendering of stochastic models.
\newblock \emph{Communications of the ACM}, 25\penalty0 (6):\penalty0 371--384, 1982.

\bibitem[Gandelsman et~al.(2022)Gandelsman, Sun, Chen, and Efros]{gandelsman2022test}
Yossi Gandelsman, Yu~Sun, Xinlei Chen, and Alexei~A Efros.
\newblock Test-time training with masked autoencoders.
\newblock \emph{arXiv preprint arXiv:2209.07522}, 2022.

\bibitem[Gao et~al.(2023)Gao, Zhang, Liu, Darrell, Shelhamer, and Wang]{Gao2023BackCorruption}
Jin Gao, Jialing Zhang, Xihui Liu, Trevor Darrell, Evan Shelhamer, and Dequan Wang.
\newblock {Back to the source: Diffusion-driven adaptation to test-time corruption}.
\newblock In \emph{Proceedings of the IEEE/CVF Conference on Computer Vision and Pattern Recognition}, pp.\  11786--11796, 2023.

\bibitem[Goodfellow et~al.(2014)Goodfellow, Shlens, and Szegedy]{goodfellow2014explaining}
Ian~J Goodfellow, Jonathon Shlens, and Christian Szegedy.
\newblock Explaining and harnessing adversarial examples.
\newblock \emph{arXiv preprint arXiv:1412.6572}, 2014.

\bibitem[Grandvalet \& Bengio(2004)Grandvalet and Bengio]{grandvalet2004semi}
Yves Grandvalet and Yoshua Bengio.
\newblock Semi-supervised learning by entropy minimization.
\newblock \emph{Advances in neural information processing systems}, 17, 2004.

\bibitem[Hendrycks \& Dietterich(2019)Hendrycks and Dietterich]{hendrycks2019robustness}
Dan Hendrycks and Thomas Dietterich.
\newblock Benchmarking neural network robustness to common corruptions and perturbations.
\newblock \emph{Proceedings of the International Conference on Learning Representations}, 2019.

\bibitem[Hu et~al.(2019)Hu, Fu, Zhu, and Heng]{Hu2019Depth-attentionalRemoval}
Xiaowei Hu, Chi-Wing Fu, Lei Zhu, and Pheng-Ann Heng.
\newblock {Depth-attentional features for single-image rain removal}.
\newblock In \emph{Proceedings of the IEEE/CVF Conference on computer vision and pattern recognition}, pp.\  8022--8031, 2019.

\bibitem[Ioffe \& Szegedy(2015{\natexlab{a}})Ioffe and Szegedy]{Ioffe2015BatchShift}
Sergey Ioffe and Christian Szegedy.
\newblock {Batch normalization: Accelerating deep network training by reducing internal covariate shift}.
\newblock In \emph{International conference on machine learning}, pp.\  448--456, 2015{\natexlab{a}}.

\bibitem[Ioffe \& Szegedy(2015{\natexlab{b}})Ioffe and Szegedy]{ioffe2015batch}
Sergey Ioffe and Christian Szegedy.
\newblock Batch normalization: Accelerating deep network training by reducing internal covariate shift.
\newblock In \emph{International conference on machine learning}, pp.\  448--456. pmlr, 2015{\natexlab{b}}.

\bibitem[Karani et~al.(2021)Karani, Erdil, Chaitanya, and Konukoglu]{karani2021test}
Neerav Karani, Ertunc Erdil, Krishna Chaitanya, and Ender Konukoglu.
\newblock Test-time adaptable neural networks for robust medical image segmentation.
\newblock \emph{Medical Image Analysis}, 68:\penalty0 101907, 2021.

\bibitem[Khurana et~al.(2021)Khurana, Paul, Rai, Biswas, and Aggarwal]{Khurana2021SITA:Adaptation}
Ansh Khurana, Sujoy Paul, Piyush Rai, Soma Biswas, and Gaurav Aggarwal.
\newblock {SITA: Single Image Test-time Adaptation}.
\newblock 12 2021.
\newblock URL \url{https://arxiv.org/abs/2112.02355v3}.

\bibitem[Kurakin et~al.(2018)Kurakin, Goodfellow, and Bengio]{kurakin2018adversarial}
Alexey Kurakin, Ian~J Goodfellow, and Samy Bengio.
\newblock Adversarial examples in the physical world.
\newblock In \emph{Artificial intelligence safety and security}, pp.\  99--112. Chapman and Hall/CRC, 2018.

\bibitem[Lee et~al.(2013)]{lee2013pseudo}
Dong-Hyun Lee et~al.
\newblock Pseudo-label: The simple and efficient semi-supervised learning method for deep neural networks.
\newblock In \emph{Workshop on challenges in representation learning, ICML}, volume~3, pp.\  896, 2013.

\bibitem[Li et~al.(2020)Li, Wang, Che, Zhang, Zhao, Xu, Zhou, Bengio, and Keutzer]{li2020rethinking}
Bo~Li, Yezhen Wang, Tong Che, Shanghang Zhang, Sicheng Zhao, Pengfei Xu, Wei Zhou, Yoshua Bengio, and Kurt Keutzer.
\newblock Rethinking distributional matching based domain adaptation.
\newblock \emph{arXiv preprint arXiv:2006.13352}, 2020.

\bibitem[Li et~al.(2018)Li, Wang, Shi, Hou, and Liu]{li2018adaptive}
Yanghao Li, Naiyan Wang, Jianping Shi, Xiaodi Hou, and Jiaying Liu.
\newblock Adaptive batch normalization for practical domain adaptation.
\newblock \emph{Pattern Recognition}, 80:\penalty0 109--117, 2018.

\bibitem[Lin et~al.(2014)Lin, Maire, Belongie, Hays, Perona, Ramanan, Doll{\'a}r, and Zitnick]{lin2014microsoft}
Tsung-Yi Lin, Michael Maire, Serge Belongie, James Hays, Pietro Perona, Deva Ramanan, Piotr Doll{\'a}r, and C~Lawrence Zitnick.
\newblock Microsoft coco: Common objects in context.
\newblock In \emph{Computer Vision--ECCV 2014: 13th European Conference, Zurich, Switzerland, September 6-12, 2014, Proceedings, Part V 13}, pp.\  740--755. Springer, 2014.

\bibitem[Liu et~al.(2021{\natexlab{a}})Liu, Zhang, and Wang]{liu2021source}
Yuang Liu, Wei Zhang, and Jun Wang.
\newblock Source-free domain adaptation for semantic segmentation.
\newblock In \emph{Proceedings of the IEEE/CVF Conference on Computer Vision and Pattern Recognition}, pp.\  1215--1224, 2021{\natexlab{a}}.

\bibitem[Liu et~al.(2021{\natexlab{b}})Liu, Kothari, van Delft, Bellot-Gurlet, Mordan, and Alahi]{liu2021ttt++}
Yuejiang Liu, Parth Kothari, Bastien van Delft, Baptiste Bellot-Gurlet, Taylor Mordan, and Alexandre Alahi.
\newblock Ttt++: When does self-supervised test-time training fail or thrive?
\newblock \emph{Advances in Neural Information Processing Systems}, 34:\penalty0 21808--21820, 2021{\natexlab{b}}.

\bibitem[Loshchilov \& Hutter(2017)Loshchilov and Hutter]{Loshchilov2017DecoupledRegularization}
Ilya Loshchilov and Frank Hutter.
\newblock {Decoupled weight decay regularization}.
\newblock \emph{arXiv preprint arXiv:1711.05101}, 2017.

\bibitem[Madry et~al.(2017)Madry, Makelov, Schmidt, Tsipras, and Vladu]{madry2017towards}
Aleksander Madry, Aleksandar Makelov, Ludwig Schmidt, Dimitris Tsipras, and Adrian Vladu.
\newblock Towards deep learning models resistant to adversarial attacks.
\newblock \emph{arXiv preprint arXiv:1706.06083}, 2017.

\bibitem[Milletari et~al.(2016)Milletari, Navab, and Ahmadi]{milletari2016dice}
Fausto Milletari, Nassir Navab, and Seyed-Ahmad Ahmadi.
\newblock V-net: Fully convolutional neural networks for volumetric medical image segmentation.
\newblock In \emph{2016 fourth international conference on 3D vision (3DV)}, pp.\  565--571. Ieee, 2016.

\bibitem[Mounsaveng et~al.(2024)Mounsaveng, Chiaroni, Boudiaf, Pedersoli, and Ben~Ayed]{mounsaveng2024bag}
Saypraseuth Mounsaveng, Florent Chiaroni, Malik Boudiaf, Marco Pedersoli, and Ismail Ben~Ayed.
\newblock Bag of tricks for fully test-time adaptation.
\newblock In \emph{Proceedings of the IEEE/CVF Winter Conference on Applications of Computer Vision}, pp.\  1936--1945, 2024.

\bibitem[Nado et~al.(2020)Nado, Padhy, Sculley, D'Amour, Lakshminarayanan, and Snoek]{nado2020evaluating}
Zachary Nado, Shreyas Padhy, D~Sculley, Alexander D'Amour, Balaji Lakshminarayanan, and Jasper Snoek.
\newblock Evaluating prediction-time batch normalization for robustness under covariate shift.
\newblock \emph{arXiv preprint arXiv:2006.10963}, 2020.

\bibitem[Nguyen et~al.({\natexlab{a}})Nguyen, Nguyen-Tang, Lim, and Torr]{NguyenTIPI:Invariance}
A~Tuan Nguyen, Thanh Nguyen-Tang, Ser-Nam Lim, and Philip H~S Torr.
\newblock {TIPI: Test Time Adaptation with Transformation Invariance}.
\newblock {\natexlab{a}}.

\bibitem[Nguyen et~al.({\natexlab{b}})Nguyen, Nguyen-Tang, Lim, and Torr]{nguyentipi}
A~Tuan Nguyen, Thanh Nguyen-Tang, Ser-Nam Lim, and Philip~HS Torr.
\newblock Tipi: Test time adaptation with transformation invariance.
\newblock {\natexlab{b}}.

\bibitem[Niu et~al.(2023)Niu, Wu, Zhang, Wen, Chen, Zhao, and Tan]{niu2023towards}
Shuaicheng Niu, Jiaxiang Wu, Yifan Zhang, Zhiquan Wen, Yaofo Chen, Peilin Zhao, and Mingkui Tan.
\newblock Towards stable test-time adaptation in dynamic wild world.
\newblock \emph{arXiv preprint arXiv:2302.12400}, 2023.

\bibitem[Prabhu et~al.(2021)Prabhu, Khare, Kartik, and Hoffman]{prabhu2021augco}
Viraj Prabhu, Shivam Khare, Deeksha Kartik, and Judy Hoffman.
\newblock Augco: augmentation consistency-guided self-training for source-free domain adaptive semantic segmentation.
\newblock \emph{arXiv preprint arXiv:2107.10140}, 2021.

\bibitem[Richter et~al.(2016)Richter, Vineet, Roth, and Koltun]{richter2016playing}
Stephan~R Richter, Vibhav Vineet, Stefan Roth, and Vladlen Koltun.
\newblock Playing for data: Ground truth from computer games.
\newblock In \emph{Computer Vision--ECCV 2016: 14th European Conference, Amsterdam, The Netherlands, October 11-14, 2016, Proceedings, Part II 14}, pp.\  102--118. Springer, 2016.

\bibitem[Rodriguez \& Mikolajczyk(2019)Rodriguez and Mikolajczyk]{rodriguez2019domain}
Adrian~Lopez Rodriguez and Krystian Mikolajczyk.
\newblock Domain adaptation for object detection via style consistency.
\newblock \emph{arXiv preprint arXiv:1911.10033}, 2019.

\bibitem[Ronneberger et~al.(2015)Ronneberger, Fischer, and Brox]{ronneberger2015unet}
Olaf Ronneberger, Philipp Fischer, and Thomas Brox.
\newblock U-net: Convolutional networks for biomedical image segmentation.
\newblock In \emph{International Conference on Medical image computing and computer-assisted intervention}, pp.\  234--241. Springer, 2015.

\bibitem[Ros et~al.(2016)Ros, Sellart, Materzynska, Vazquez, and Lopez]{ros2016synthia}
German Ros, Laura Sellart, Joanna Materzynska, David Vazquez, and Antonio~M Lopez.
\newblock The synthia dataset: A large collection of synthetic images for semantic segmentation of urban scenes.
\newblock In \emph{Proceedings of the IEEE conference on computer vision and pattern recognition}, pp.\  3234--3243, 2016.

\bibitem[Saito et~al.(2019)Saito, Kim, Sclaroff, Darrell, and Saenko]{Saito2019Semi-supervisedEntropy}
Kuniaki Saito, Donghyun Kim, Stan Sclaroff, Trevor Darrell, and Kate Saenko.
\newblock {Semi-supervised domain adaptation via minimax entropy}.
\newblock In \emph{Proceedings of the IEEE/CVF international conference on computer vision}, pp.\  8050--8058, 2019.

\bibitem[Schneider et~al.(2020)Schneider, Rusak, Eck, Bringmann, Brendel, and Bethge]{schneider2020improving}
Steffen Schneider, Evgenia Rusak, Luisa Eck, Oliver Bringmann, Wieland Brendel, and Matthias Bethge.
\newblock Improving robustness against common corruptions by covariate shift adaptation.
\newblock \emph{Advances in Neural Information Processing Systems}, 33:\penalty0 11539--11551, 2020.

\bibitem[Steiner et~al.(2021)Steiner, Kolesnikov, Zhai, Wightman, Uszkoreit, and Beyer]{steiner2021train}
Andreas Steiner, Alexander Kolesnikov, Xiaohua Zhai, Ross Wightman, Jakob Uszkoreit, and Lucas Beyer.
\newblock How to train your vit? data, augmentation, and regularization in vision transformers.
\newblock \emph{arXiv preprint arXiv:2106.10270}, 2021.

\bibitem[Sun et~al.(2020)Sun, Wang, Liu, Miller, Efros, and Hardt]{sun2020test}
Yu~Sun, Xiaolong Wang, Zhuang Liu, John Miller, Alexei Efros, and Moritz Hardt.
\newblock Test-time training with self-supervision for generalization under distribution shifts.
\newblock In \emph{International conference on machine learning}, pp.\  9229--9248. PMLR, 2020.

\bibitem[Tan \& Le(2019)Tan and Le]{tan2019efficientnet}
Mingxing Tan and Quoc Le.
\newblock Efficientnet: Rethinking model scaling for convolutional neural networks.
\newblock In \emph{International conference on machine learning}, pp.\  6105--6114. PMLR, 2019.

\bibitem[Tanwisuth et~al.(2021)Tanwisuth, Fan, Zheng, Zhang, Zhang, Chen, and Zhou]{tanwisuth2021prototype}
Korawat Tanwisuth, Xinjie Fan, Huangjie Zheng, Shujian Zhang, Hao Zhang, Bo~Chen, and Mingyuan Zhou.
\newblock A prototype-oriented framework for unsupervised domain adaptation.
\newblock \emph{Advances in Neural Information Processing Systems}, 34:\penalty0 17194--17208, 2021.

\bibitem[Tzeng et~al.(2017)Tzeng, Hoffman, Saenko, and Darrell]{tzeng2017adversarial}
Eric Tzeng, Judy Hoffman, Kate Saenko, and Trevor Darrell.
\newblock Adversarial discriminative domain adaptation.
\newblock In \emph{Proceedings of the IEEE conference on computer vision and pattern recognition}, pp.\  7167--7176, 2017.

\bibitem[Valvano et~al.(2021)Valvano, Leo, and Tsaftaris]{valvano2021re}
Gabriele Valvano, Andrea Leo, and Sotirios~A Tsaftaris.
\newblock Re-using adversarial mask discriminators for test-time training under distribution shifts.
\newblock \emph{arXiv preprint arXiv:2108.11926}, 2021.

\bibitem[Volpi et~al.(2022)Volpi, De~Jorge, Larlus, and Csurka]{volpi2022road}
Riccardo Volpi, Pau De~Jorge, Diane Larlus, and Gabriela Csurka.
\newblock On the road to online adaptation for semantic image segmentation.
\newblock In \emph{Proceedings of the IEEE/CVF Conference on Computer Vision and Pattern Recognition}, pp.\  19184--19195, 2022.

\bibitem[Wang et~al.(2020{\natexlab{a}})Wang, Shelhamer, Liu, Olshausen, and Darrell]{Wang2020Tent:Minimization}
Dequan Wang, Evan Shelhamer, Shaoteng Liu, Bruno Olshausen, and Trevor Darrell.
\newblock {Tent: Fully test-time adaptation by entropy minimization}.
\newblock \emph{arXiv preprint arXiv:2006.10726}, 2020{\natexlab{a}}.

\bibitem[Wang et~al.(2020{\natexlab{b}})Wang, Shelhamer, Liu, Olshausen, and Darrell]{wang2020tent}
Dequan Wang, Evan Shelhamer, Shaoteng Liu, Bruno Olshausen, and Trevor Darrell.
\newblock Tent: Fully test-time adaptation by entropy minimization.
\newblock \emph{arXiv preprint arXiv:2006.10726}, 2020{\natexlab{b}}.

\bibitem[Wang et~al.(2022)Wang, Fink, Van~Gool, and Dai]{wang2022continual}
Qin Wang, Olga Fink, Luc Van~Gool, and Dengxin Dai.
\newblock Continual test-time domain adaptation.
\newblock In \emph{Proceedings of the IEEE/CVF Conference on Computer Vision and Pattern Recognition}, pp.\  7201--7211, 2022.

\bibitem[Wang et~al.()Wang, Sun, Gandelsman, Chen, Efros, and Wang]{WangTest-TimeStreams}
Renhao Wang, Yu~Sun, Yossi Gandelsman, Xinlei Chen, Alexei~A Efros, and Xiaolong Wang.
\newblock {Test-Time Training on Video Streams}.
\newblock URL \url{https://video-ttt.github.io/}.

\bibitem[Wightman(2019)]{rw2019timm}
Ross Wightman.
\newblock Pytorch image models.
\newblock \url{https://github.com/rwightman/pytorch-image-models}, 2019.

\bibitem[Yi et~al.(2023)Yi, Chen, and Zhang]{yi2023critical}
Chang'an Yi, Haotian Chen, and Yifan Zhang.
\newblock A critical look at classic test-time adaptation methods in semantic segmentation.
\newblock \emph{arXiv preprint arXiv:2310.05341}, 2023.

\bibitem[Yu et~al.(2023)Yu, Sheng, He, and Liang]{yu2023benchmarking}
Yongcan Yu, Lijun Sheng, Ran He, and Jian Liang.
\newblock Benchmarking test-time adaptation against distribution shifts in image classification.
\newblock \emph{arXiv preprint arXiv:2307.03133}, 2023.

\bibitem[Zhang et~al.(2018)Zhang, Miao, Mansi, and Liao]{zhang2018task}
Yue Zhang, Shun Miao, Tommaso Mansi, and Rui Liao.
\newblock Task driven generative modeling for unsupervised domain adaptation: Application to x-ray image segmentation.
\newblock In \emph{Medical Image Computing and Computer Assisted Intervention--MICCAI 2018: 21st International Conference, Granada, Spain, September 16-20, 2018, Proceedings, Part II}, pp.\  599--607. Springer, 2018.

\end{thebibliography}
\bibliographystyle{tmlr}
\clearpage
\appendix
\section{Related Domain Adaptation Scenarios}

\label{sec:ap:rel}

\textbf{Continuous domain adaptation} assumes the target data change continually, as opposed to a static target distribution. Furthermore, it is important to avoid catastrophic forgetting (previous knowledge). In this work, we make no assumption about the relationship between the distribution of subsequent samples, each sample could come from a different distribution. For this reason, our models are initialized to the pretrained weights before adapting to a new image and catastrophic forgetting is not a concern. While single-image adaptation methods can be extended to continual learning, it is not clear that methods performing better in the single-image setup will also perform better in batch or stream of data mode.

\textbf{Domain generalization} aims for a strong model that would generalize to unseen domains without any adaptation. Common approaches are domain-invariant representation learning, data augmentation or data generation. In contrast, this work focuses on adapting a pre-trained model to become a specialist in the current domain. It was shown in \cite{volpi2022road} that a stronger, more general model can lead to better adaptation results, making these directions complementary. However, without adaptation, training a very general model on a large set of distributions may harm the model's performance on the individual distributions when compared to specialized models with carefully optimized data augmentation \cite{aquino2017effect, steiner2021train}.

\textbf{Online domain adaptation} expects a stream of data as input, possibly a single one, as opposed to a whole dataset. Test-time adaptation methods can be sued for online adaptation but generally, online adaptation techniques do not assume the source data are not available.

\textbf{Zero-shot segmentation} requires the model to directly perform predictions for previously unseen classes without any adaptation to these classes.
\section{Baseline methods} \label{sec:ap:baselines}
In this section, the self-supervised loss functions optimized in the baseline methods are described, as well as the adaptations from the original implementation to the single-image setup when necessary.

\subsection{Entropy Minimization (Ent)}
The \ac{ttaent} method minimizes the entropy of the segmentation predictions. In the context of learning with limited supervision, it was proposed in \cite{grandvalet2004semi} for semi-supervised learning. In \ac{tta}, there is no labelled set that could be leveraged as regularization like in semi-supervised learning but the methods were shown to work for \ac{tta} as well \cite{wang2020tent}.
It was also shown that larger batch size and updating the parameters of the normalization-layers only improve stability of the method. But on segmentation, a dense-prediction task, adapting to a single image can lead to positive results, especially when not updating the batch normalization statistics but only the learnable parameters like in \cite{volpi2022road}.

The method is simple, computationally efficient and widely adopted as a baseline.
More formally, the method minimizes the entropy of the segmentation model predictions $s = f^{\theta}_\text{S}(x)$ for an image $x$:

\begin{equation}
    \mathcal{L}_{\text{Ent}} = \sum_{i=1}^\text{N}\sum_{i=1}^{\text{C}} \text{s}_{ic} \cdot \log(\text{s}_{ic})
\end{equation}

where $\text{C}$ is the total number of classes, $\text{s}_{ic}$ corresponds to the $i$-th pixel of the segmentation prediction $\text{s}$ for class $c$ and $\text{N}$ is the total number of pixels in the image.

 In this work, the batch normalization \cite{ioffe2015batch} statistics are not updated since it relies on the presence of batch normalization layers while many recent architectures use other normalization layers such as layer normalization \cite{ba2016layer}.

\subsection{Pseudolabelling (PL)}
Pseudolabelling also comes from semi-supervised learning \cite{lee2013pseudo} and is based on the idea of using the segmentation prediction of an image (prediction for each class binarized through argmax) as ground truth to optimize the model. In effect, it is the same as pseudo-labelling since both methods reduce class overlap. However, pseudolabelling has the advantage of allowing for different loss function. While the \ac{ce} loss is typically optimized, our experiments show that in the single image setup, \ac{iou} leads to superior results.

\subsection{Adversarial Transformation Invariance (Adv)}
This method is an extension of TIPI (Test-Time Adaptation with Transformation Invariance) by \cite{nguyentipi} to image segmentation. 
The main idea is to make the network invariant to adversarial transformation of the input image as a representative of domain shifts.

The optimization loss is computed as the reverse KL divergence loss between the model prediction $s' = f^{\theta}_\text{S}(x')$ where $x'$ is an adversarially transformed image and the prediction on the original input $s = f^{\theta}_\text{S}(x)$.

\begin{equation}
     \mathcal{L}_\text{Adv} = \mathcal{L}_\text{KL}(\text{s}_i, \text{s}'_i)
\end{equation}

where $\text{s}'_i$ is the adversarially transformed prediction and KL is the Kullback-Leibler divergence loss defined as

\begin{equation}
    \mathcal{L}_\text{KL}(\text{p}, \text{q}) = 
    \frac{1}{N}
    \sum_{i=1}^N  
    \text{q}_i \cdot \log (\frac{\text{q}_i}{\text{p}_i} )
\end{equation}

In forward KL, $\text{p}$ corresponds to the model prediction while $\text{q}$ to the ground truth. Please note that, as suggested in \cite{nguyentipi}, the reverse KL is used in the proposed method where the input arguments to the function are switched, compared to forward KL.
Another important implementation detail is that the gradients should not flow through $\text{s}'_i$ - the tensor needs to be detached before the loss computation.

The same adversarial attacks in terms of the ground truth as for the IoU estimation and mask refinement methods are used to generate $x'$ but the computational complexity is reduced by using the Fast Gradient Sign Attack (FGSM) proposed in \cite{goodfellow2014explaining}.

Importantly, two sets of batch normalization statistics are kept in \cite{nguyentipi}, which is not done in our work due to the aim for general methods that do nto assuem the presence of specific network layers. This may be the reason why the methods perform poorly in our experiments. Another thing we ntoed is the high variance of the KL loss.

\subsection{Augmentation Consistency (AugCo)}
\begin{figure}[bt]
    \centering
        \includegraphics[keepaspectratio, width=0.8\linewidth]{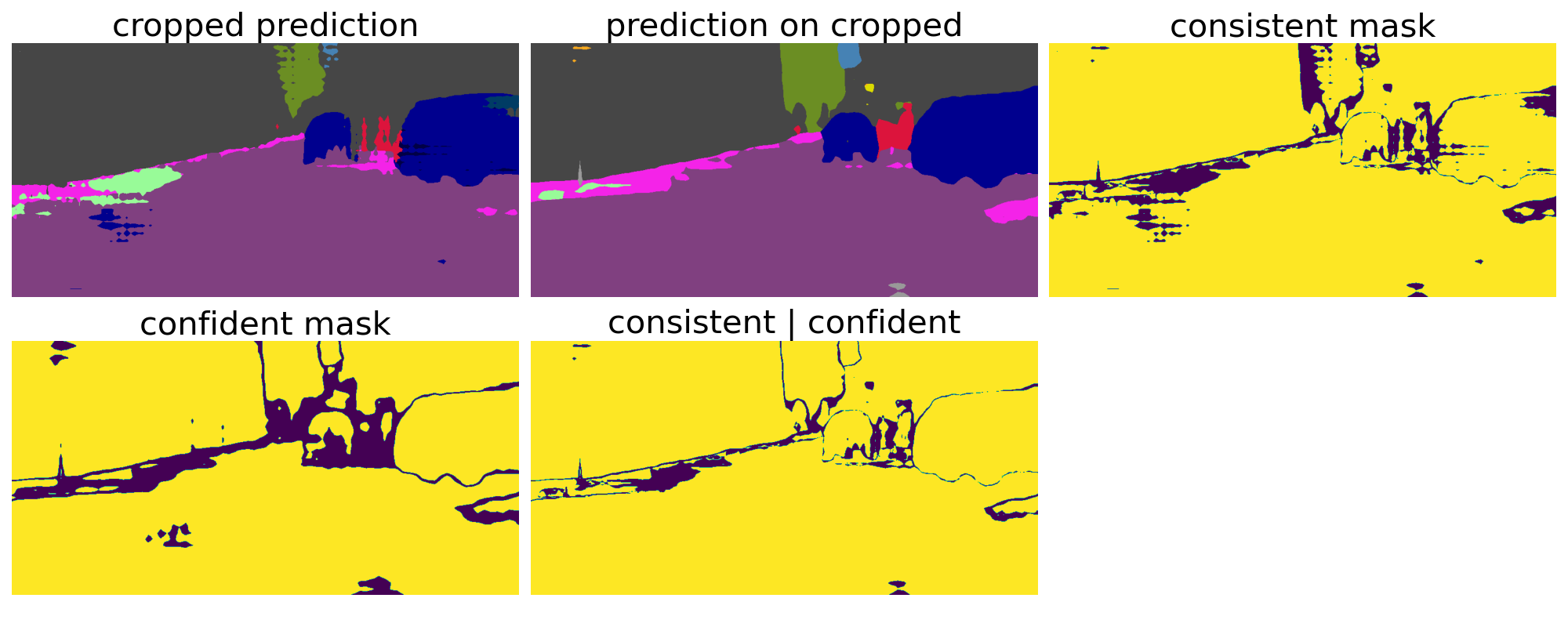}
    \caption{Visualization of the predicted views, confidence, consistency and reliablity (confident $|$ reliable) masks used by the \ac{ttaaugco} \ac{tta} method. Confident/consistent/reliable predictions are shown in yellow.}
    \label{fig:gta5:augco}
\end{figure}
The method of \cite{prabhu2021augco} is adapted to the single image setup. 
The idea is to create two segmentation views based on the input image, both based on a random bounding box with parameters $\alpha$. The bounding box should take 25-50 \% of the original image area and preserve its aspect ratio. 
View 1 is created by cropping and resizing the segmentation of the original image, $V_1 = \text{resize}(\text{crop}_\alpha(s), \text{H}, \text{W})$ where $s = f^{\theta}_\text{S}(x)$ is the segmentation prediction for an image $x$ of spatial dimensions $\text{H}, \text{W}$. 
View 2 is created as the segmentation prediction on a cropped, resized and randomly augmented image,  $V_2 = f^{\theta}_\text{S}(x')$ where  $x' = \text{resize}(\text{crop}_\alpha(\text{jitter}(x)), \text{H}, \text{W})$. 

Finally, two masks are created to identify reliable predictions: Consistency mask based on consistency between the predictions of the two views, and confidence mask based on the confidence in the prediction in $V_2$, binarized with a confidence threshold $\theta$. These are then combined with the $\mathrm{OR}$ operation. We set $\theta = 0.8$. 

The network parameters are then updated via pseudo-labelling based on predictions of $V_2$ and using reliable pixels only. 

In \cite{prabhu2021augco}, an auxiliary information entropy loss preventing trivial solutions is also optimized. This loss requires running class-frequency statistics and is not applicable to the single-image setup. Further, an adaptive threshold based on per-class confidence distirbution in a batch of images is computed iinstead of a fixed threshold, which is also not applicable in our setup. 

An example of the two views and the consistency and confidence masks, as well as the resulting reliability masks, are shown in Figure \ref{fig:gta5:augco}.
For more details please see the original work.

\section{Adversarial refinement training} \label{sec:ap:adv_ref}
\begin{figure*}[bth]
    \centering
        \includegraphics[keepaspectratio, width=\linewidth]{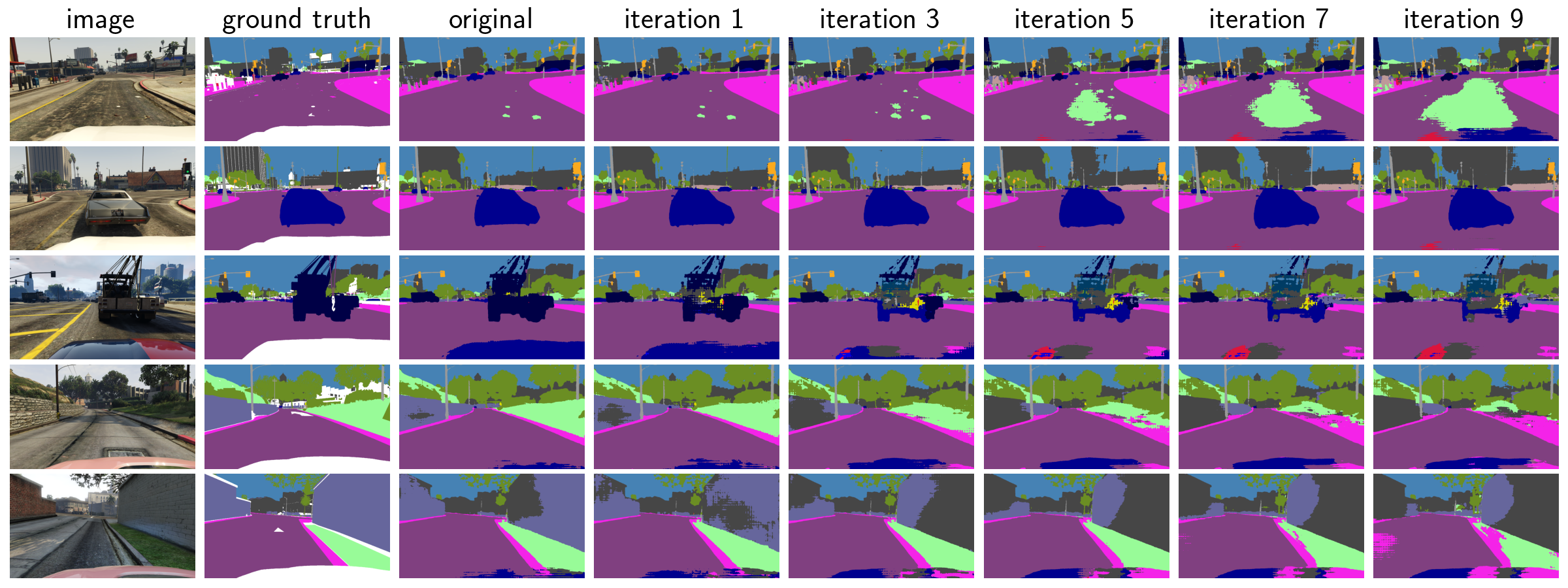}
    \caption{Evolution of masks over iterations of a projected gradient descent adversarial attack on the input image, the target being mask inversion for all of the classes. These masks serve as training data for the refinement module.}
    \label{fig:gta5:adv_mask_evol}
\end{figure*}

A visulaisation of mask evolution as the adversarial attack progresses is shown in Figure \ref{fig:gta5:adv_mask_evol}. It can be observed that the first iterations typically result in very small changes in easily confused areas, turning into more and more distorted masks.

Instead of the iterative Projected Gradient Descent (PGD). However, since the projection only consists in restricting the output to a valid range for an image, typically implemented by simply clipping the output, it is also often referred to as iterative FGSM.

\section{Synthetic corruptions} \label{sec:ap:corrs}
The corruptions used in our experiments are a subset of the corruptions from \cite{hendrycks2019robustness}.
An overview of the corruptions, as well as implementation details, can be found in Table \ref{tab:distortion_def}

\begin{table*}
{
\renewcommand{\arraystretch}{1.5}
    \begin{tabular}{p{0.18\textwidth}  p{0.75\textwidth} }
    \toprule

    corruption & description \\

    \midrule
    %     \multicolumn{2}{l}{intensity corruptions} \\
    % \midrule
    \textbf{brightness} & is an additive intensity transformation, $x_c = \text{clip}(x + b)$ where $b$ controls the level. \\
    
    \textbf{contrast} & is a multiplicative intensity transformation, $x_c = b(x - \overline{x}) + x$ where $b$ controls the level and $\overline{x}$ is a per-channel mean of the image intensities. \\

    % \midrule
    % \multicolumn{2}{l}{weather corruptions} \\
    % \midrule
    
    \textbf{frost} & first crops a portion of one of the frost image templates at a random location of the same size as the input image, $x_\text{f}$. Then we compute $ x_c = b_1 x + b_2 x_\text{f}$  where the weights $b_1, b_2$ control the level. \\
    
    \textbf{fog} & first generates a heightmap $x_\text{h}$ using the diamond-square algorithm \cite{fournier1982computer}, where the wibble is controlled by a parameter $b_1$. It is then combined with the input image as $ x_c = \frac{(x + b_1 \cdot x_\text{h}) \cdot \max(x)} {\max(x) + b_1}$. \\
    % TODO denominator decreases contrast, max in nominator decreases brightness?

    % \midrule
    %     \multicolumn{2}{l}{noise corruptions} \\
    % \midrule
    
    \textbf{gaussian noise} & is generated as $ x_c =
 x + n$ 
    where $n \sim \mathcal{N} (0, b) $ and $b$ controls the level. \\
    
    \textbf{shot noise} & is generated as $ x_c \sim
  \frac{\text{Pois}(x\cdot b, \lambda=1)}{b}$ 
    where $\text{Pois}$ denotes the Poisson distribution and $b$ controls the level. \\
    
    \textbf{spatter} & simulates mud or water spoiling. The main idea of the algorithm is a combination of thresholding and blurring random noise. \\

    % \midrule
    % \multicolumn{2}{l}{blur corruptions} \\
    % \midrule
    
    \textbf{defocus blur} & first generates a disk kernel $\text{K}$ with radius $b_1$ and alias blur $b_2$. The kernel is then used to filter each of the channels $ x_c = \text{K}(x)$. \\
    
    \textbf{gaussian blur} & corrupts the image by gaussian blurring $ x_c = \mathcal{N} (x, b)$ 
    where  $b$ controls the level. \\

    \textbf{jpeg} & is computed as $x_c = \text{jpeg}(x, b)$ where $\text{jpeg}$ performs the JPEG compression with quality $b$. \\
    
    \bottomrule
    \end{tabular}

    \caption[Corruptions and their implementation details.]{Corruptions and their implementation details, a subset from \cite{hendrycks2019robustness}. The input of the transformation is an image $x$ normalized to the $(0, 1)$ range, the output is a corrupted image $x_c$. The $\text{clip}$ function limits the values to the $[0, 1]$ range. This function is always applied to the output image after the transformation to obtain the final output $x_f = \text{clip}(x_c)$. For more details on the transformations and the values defining the level, please refer to the codebase. 
}
    \label{tab:distortion_def}
}
\end{table*}

\section{Additional experimental results} \label{sec:ap:exps}

\begin{figure*}[tbh]
    \centering
        \includegraphics[keepaspectratio, width=\linewidth]{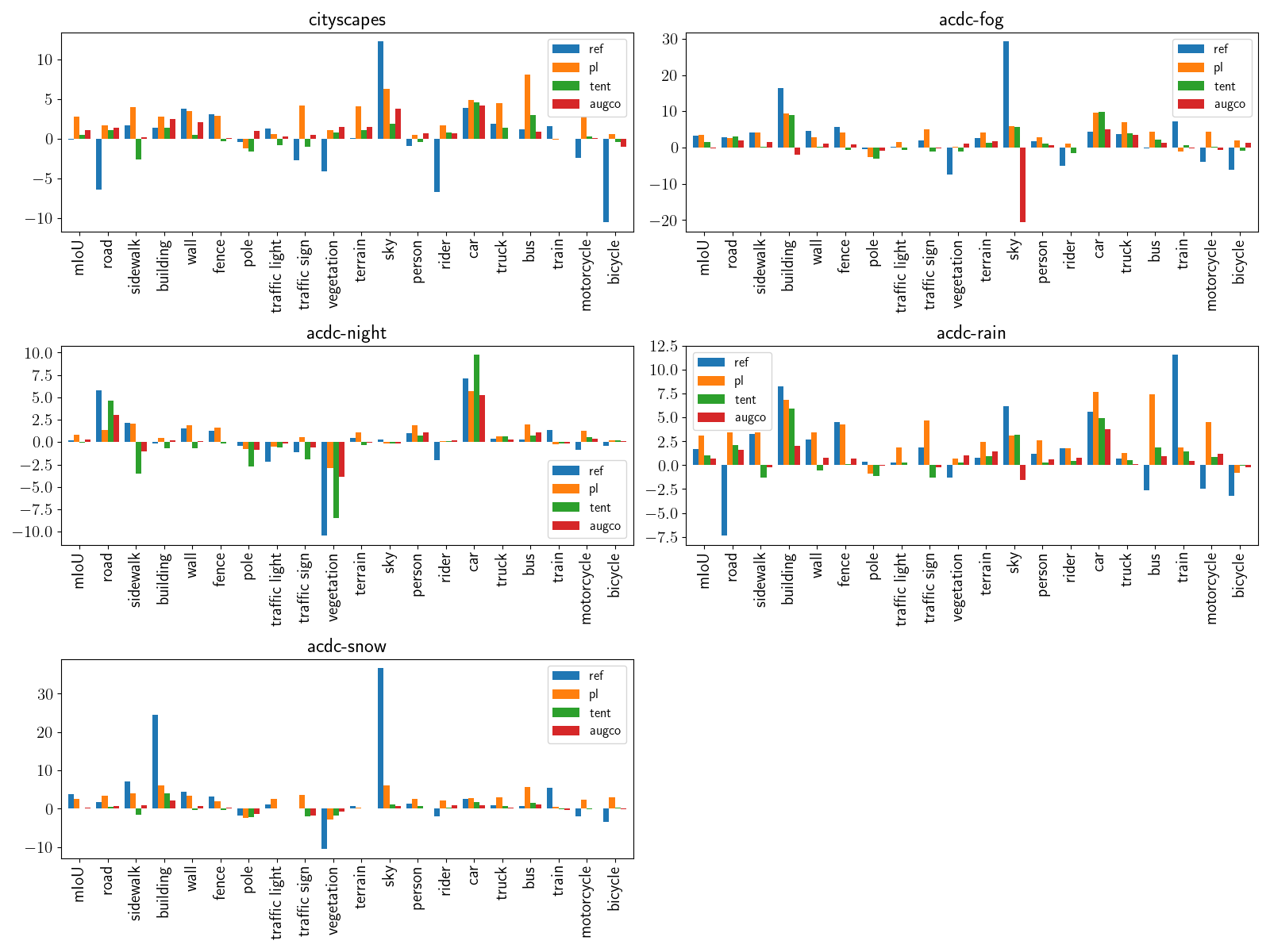}
    \caption{GTA5-C: $\overline{\text{IoU}}_c$ per-class comparison of different \ac{tta} methods.}
    \label{fig:driving:test_class_iou}
\end{figure*}

\textbf{\ac{sitta} training results}

\begin{figure*}[tbh]
    \centering
        \includegraphics[keepaspectratio, width=0.98\linewidth]{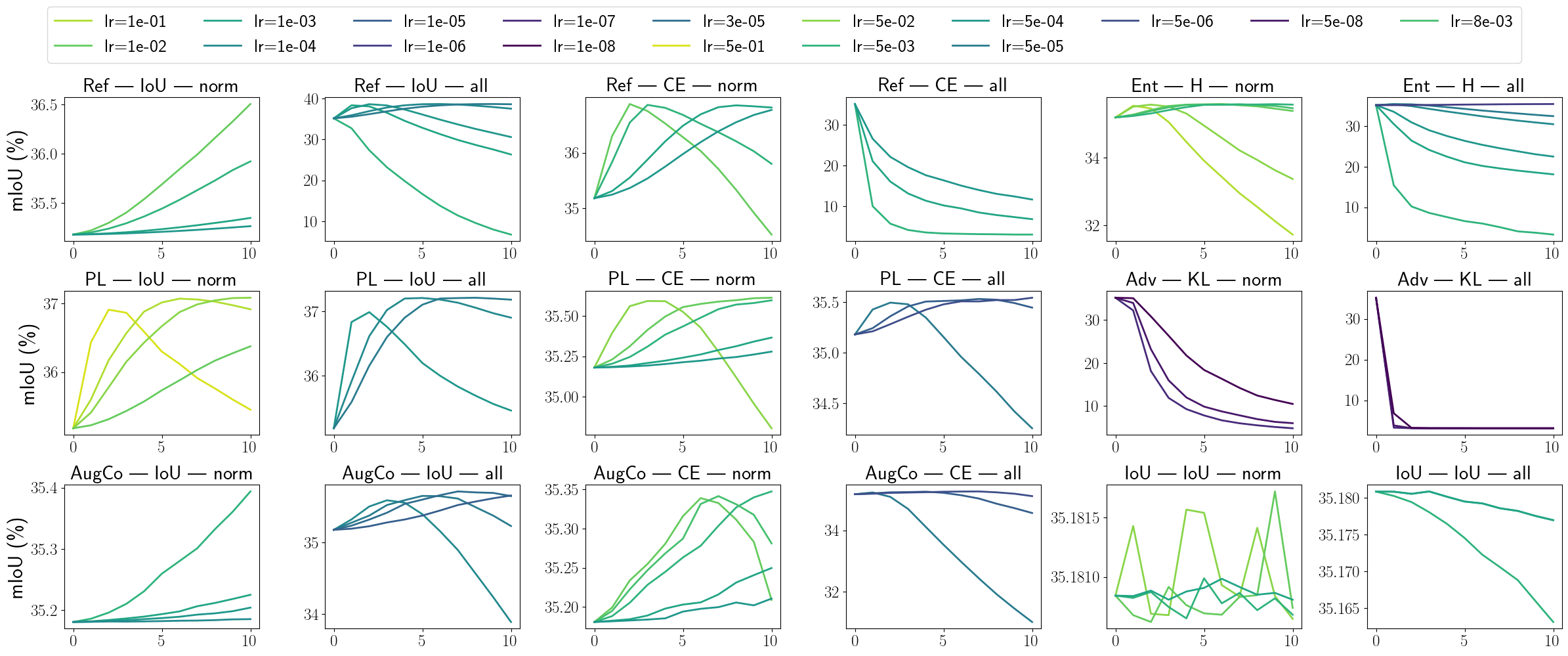}
    \caption{GTA5: $\text{m}_i\overline{\text{IoU}}$ evolution over 10 TTA iterations as a function of the learning rate. The results are reported as `method - loss - optimized parameters'. The y-axes scale differs for each subplot to better visualize learning-rate differences for each method.}
    \label{fig:gta5:lrs}
\end{figure*}

\begin{figure*} [tbh]
    \centering
        \includegraphics[keepaspectratio, width=0.98\linewidth]{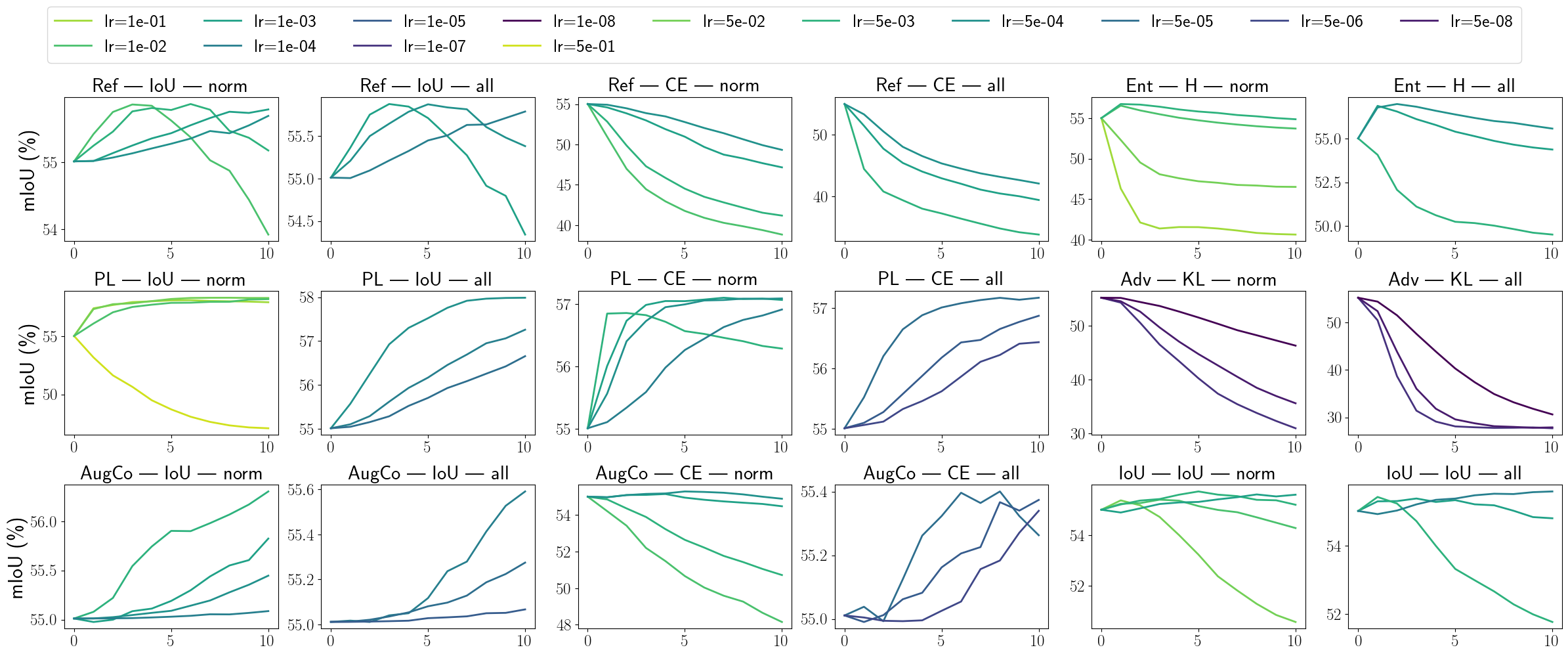}
    \caption{COCO-C: $\text{m}_i\overline{\text{IoU}}$ evolution over 10 TTA iterations as a function of the learning rate. The results are reported as 
    `method - loss - optimized parameters'. The y-axes scale differs for each subplot to better visualize learning-rate differences for each method.}
    \label{fig:coco:lrs}
\end{figure*}

\begin{table}[htb]
    \setlength{\tabcolsep}{3pt}
    % \small
    \centering
        \begin{tabular}{lccc}
        \toprule
         & & \multicolumn{2}{c}{$\text{m}\overline{\text{IoU}}_i$}  \\
            \cmidrule(lr){3-4} 
          dataset & trained on  &  non-adapted &  TTA \\
\midrule
            \multirow{2}{*}{COCO-C} & predictions & 55.01 & 57.31 \\
             & gts & 55.01 & 55.88 \\
\midrule
            \multirow{2}{*}{GTA5-C} & predictions & 35.18 & 38.63  \\
             & gts & 35.18 & 38.69  \\
\bottomrule
\end{tabular}

        \caption{Comparison of training the refinement module with ground truth masks and with segmentation model predictions. The  $\text{m}\overline{\text{IoU}}_i$ aggregated across all corruption types and levels is reported with overall optimal hyper-parameters for each dataset.
        }
        \label{tab:sem_seg:ref_gt_abl}
\end{table}
The evolution of $\text{m}\overline{\text{IoU}}_i$  over \ac{tta} iterations depending on the hyper-parameters on the GTA5-C validation set can be found in Figure \ref{fig:gta5:lrs}. The same results for the COCO-C validation dataset can be found in Figure \ref{fig:coco:lrs}.

\textbf{Test results} 
The $\text{m}\overline{\text{IoU}}_c$ comparison for all classes on the ACDC and Cityscapes test datasets can be found in Figure \ref{fig:driving:test_class_iou}.

\textbf{Additional experiments}
The refinement module is trained to predict a clean mask based on a corrupted mask simulating masks processed by the model under domain shift. The clean mask can be the segmentation prediction on clean, non-corrupted images or, when available, ground truth masks can be used instead. Comparison of these two choices is shown in Table \ref{tab:sem_seg:ref_gt_abl}.
The results are somewhat inconclusive - for the COCO model it can be seen that the model trained on predictions is substantially better than the one trained on ground truth. The GTA5 model performs similarly in both cases. One could argue that learning with GT can compensate for some of the mistakes even source distribution images, since prediction from output space back to output space is different then prediction from image to output space. On the other hand, when predictions differ significantly from ground truth even on source distirbution images, it can result in noiser data and more difficult training. The choice should be validated experimentally for each model and dataset.

\begin{figure*}[tb]
    \centering
        \includegraphics[keepaspectratio, width=0.9\linewidth]{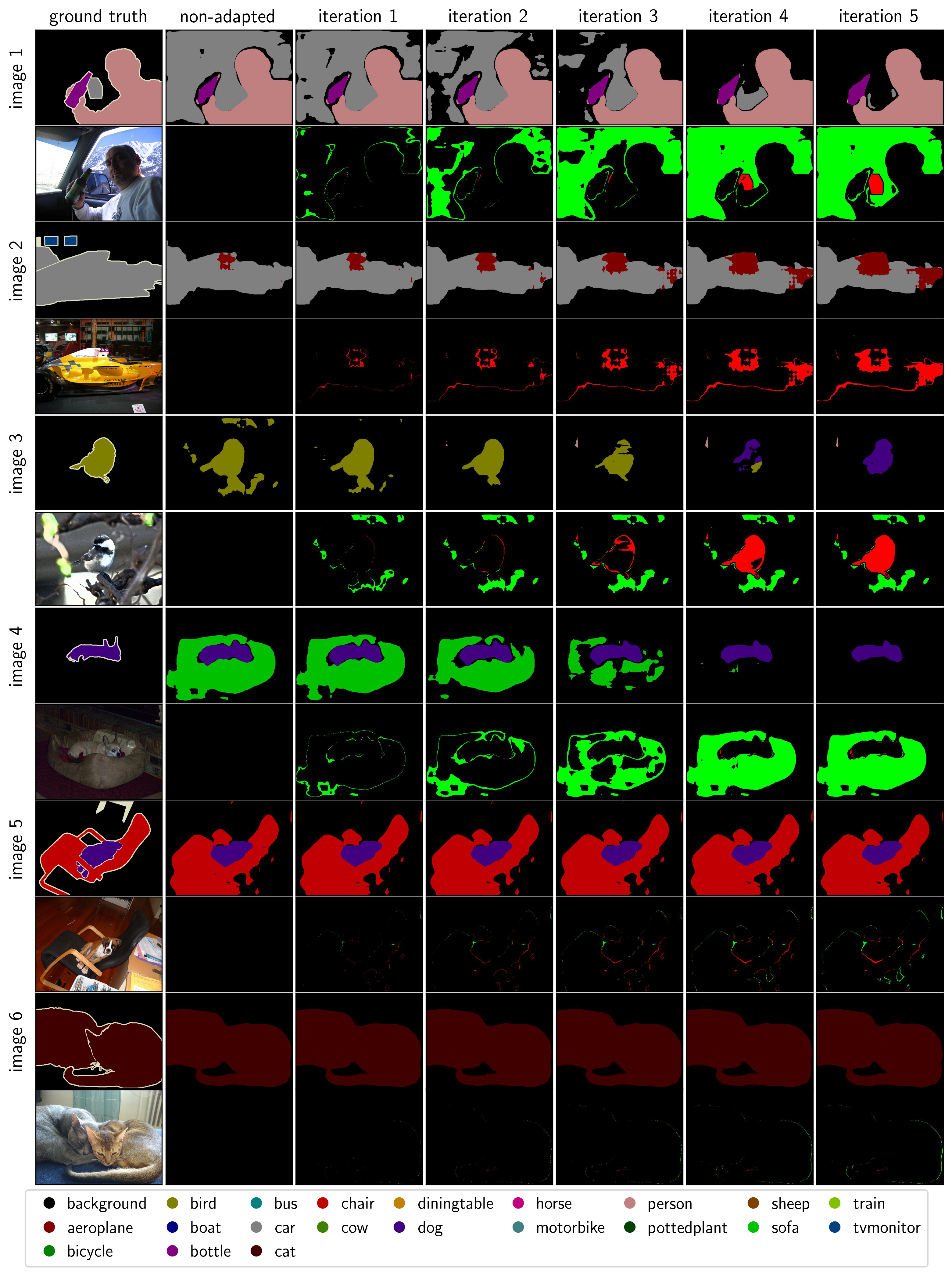}
    \caption{Segmentation evolution during \ac{tta} with \ac{ttaref} on VOC test set. First row shows the evolution of masks, second row shows the input image and segmentation improvement w.r.t. to the non-adapted mask. \textcolor{green}{Improved} and \textcolor{red}{deteriorated} pixels are highlighted.}
    \label{fig:voc_evol}
\end{figure*}

\begin{figure*}[bt]
    \centering
        \includegraphics[keepaspectratio, width=0.9\linewidth]{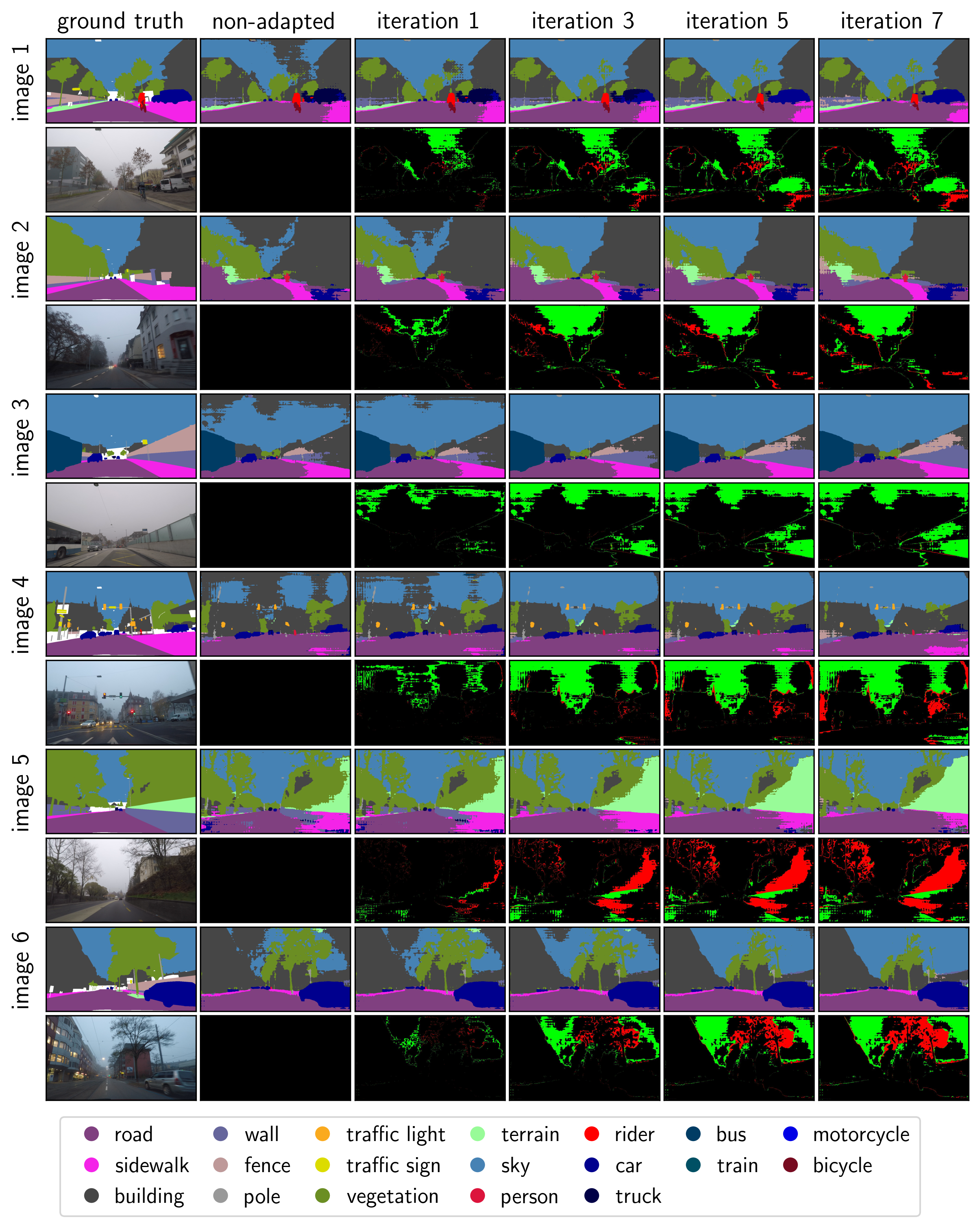}
    \caption{Segmentation evolution during \ac{tta} with \ac{ttaref} on ACDC-fog test set. First row shows the evolution of masks, second row shows the input image and segmentation improvement w.r.t. to the non-adapted mask. \textcolor{green}{Improved} and \textcolor{red}{deteriorated} pixels are highlighted.}
    \label{fig:driving:acdc_fog_evol}
\end{figure*}
\begin{figure*}[bt]
    \centering
        \includegraphics[keepaspectratio, width=0.9\linewidth]{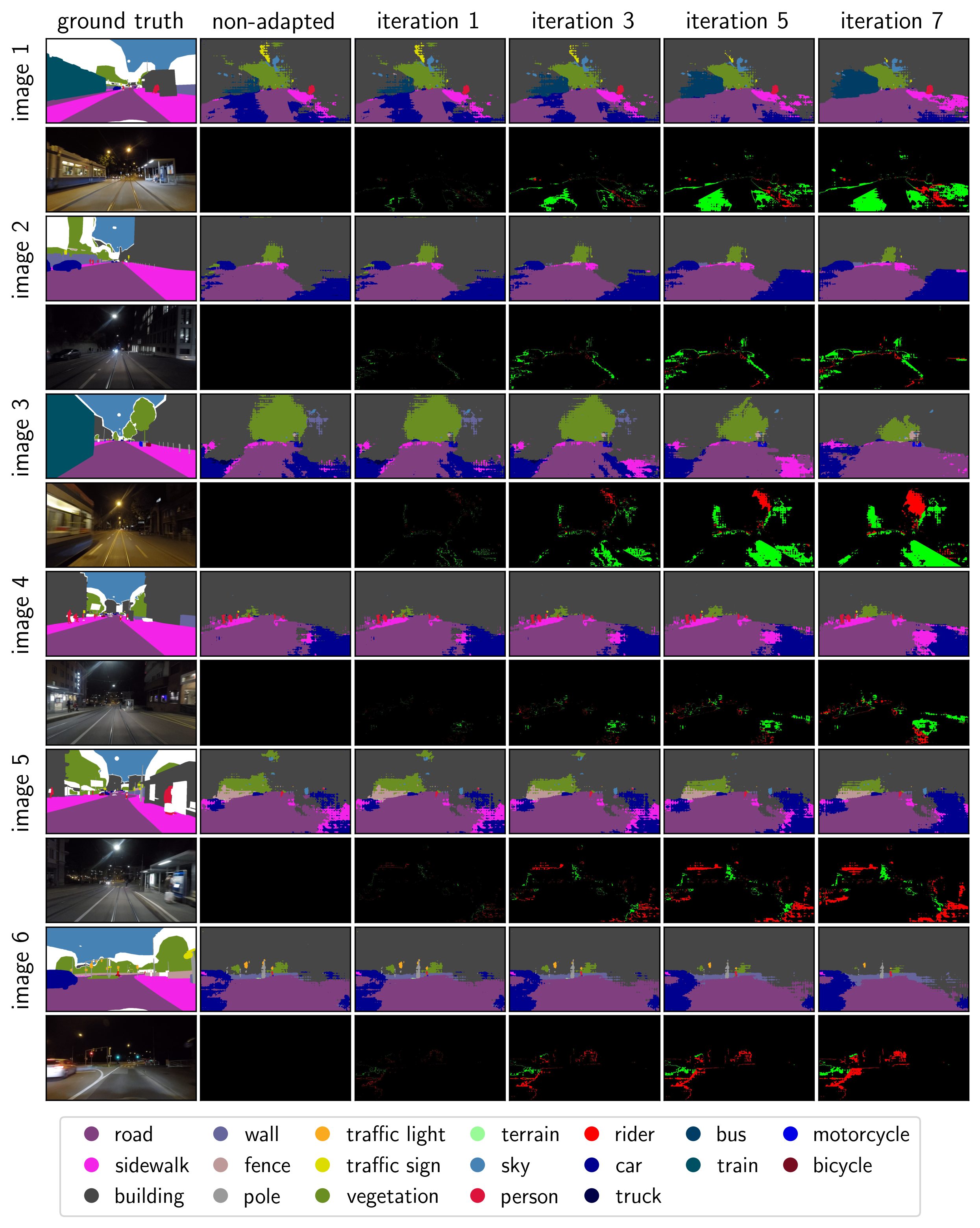}
    \caption{Segmentation evolution during \ac{tta} with \ac{ttaref} on ACDC-night test set. First row shows the evolution of masks, second row shows the input image and segmentation improvement w.r.t. to the non-adapted mask. \textcolor{green}{Improved} and \textcolor{red}{deteriorated} pixels are highlighted.}
    \label{fig:driving:acdc_night_evol}
\end{figure*}
\begin{figure*}[bt]
    \centering
        \includegraphics[keepaspectratio, width=0.9\linewidth]{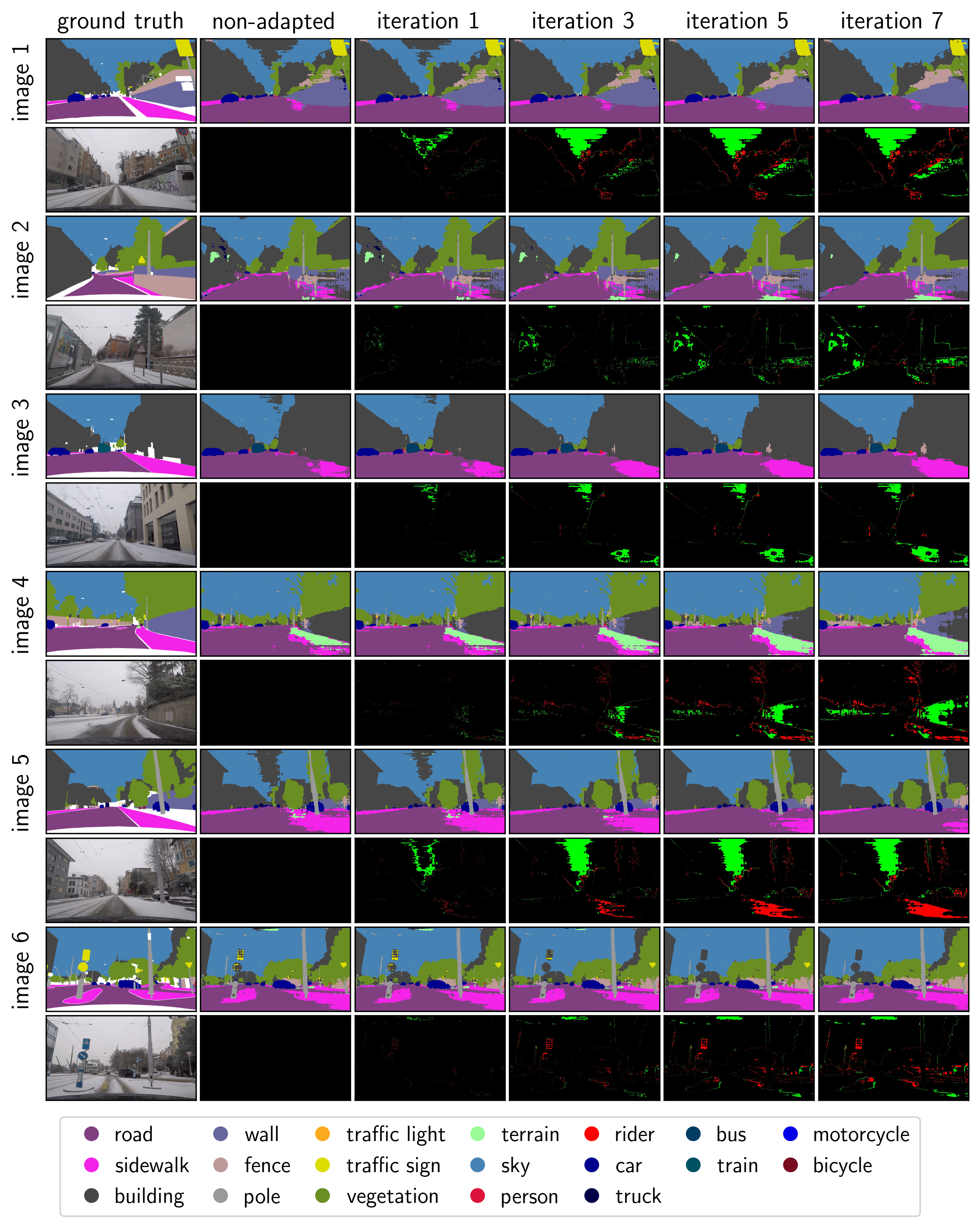}
    \caption{Segmentation evolution during \ac{tta} with \ac{ttaref} on ACDC-snow test set. First row shows the evolution of masks, second row shows the input image and segmentation improvement w.r.t. to the non-adapted mask. \textcolor{green}{Improved} and \textcolor{red}{deteriorated} pixels are highlighted.}
    \label{fig:driving:acdc_snow_evol}
\end{figure*}
\begin{figure*}[bt]
    \centering
        \includegraphics[keepaspectratio, width=0.9\linewidth]{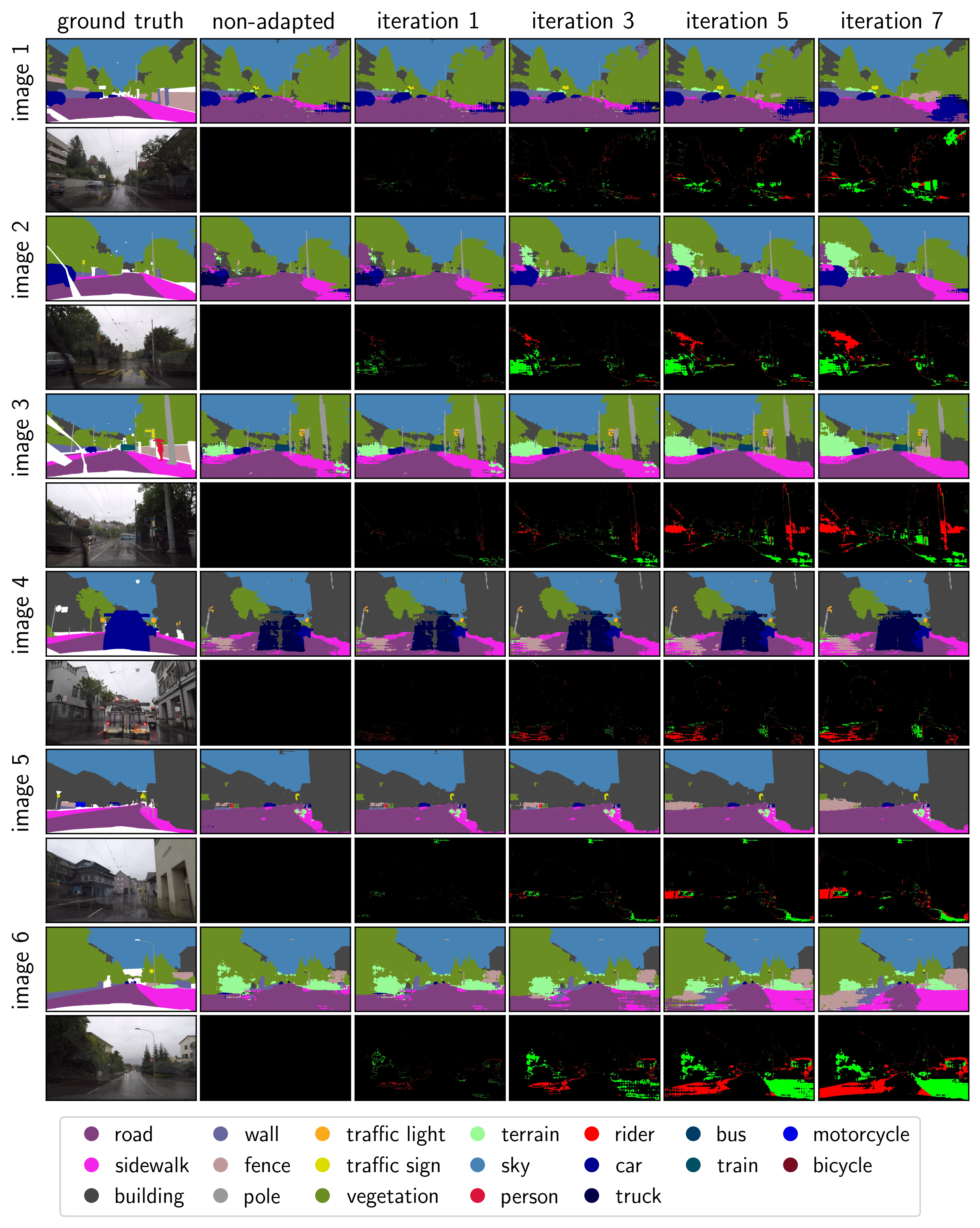}
    \caption{Segmentation evolution during \ac{tta} with \ac{ttaref} on ACDC-rain test set. First row shows the evolution of masks, second row shows the input image and segmentation improvement w.r.t. to the non-adapted mask. \textcolor{green}{Improved} and \textcolor{red}{deteriorated} pixels are highlighted.}
    \label{fig:driving:acdc_rain_evol}
\end{figure*}
\begin{figure*}[bt]
    \centering
        \includegraphics[keepaspectratio, width=0.9\linewidth]{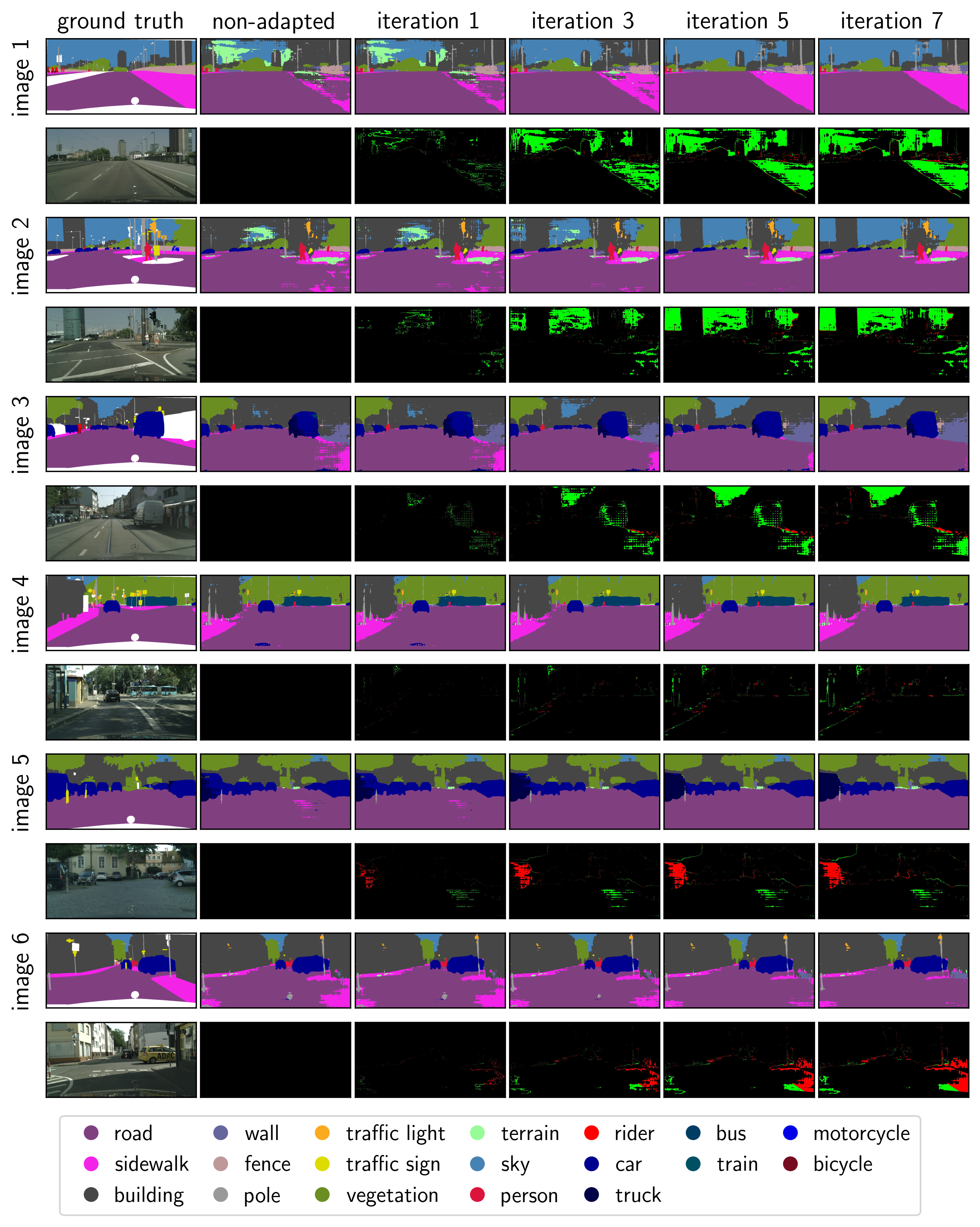}
    \caption{Segmentation evolution during \ac{tta} with \ac{ttaref} on cityscapes test set. First row shows the evolution of masks, second row shows the input image and segmentation improvement w.r.t. to the non-adapted mask. \textcolor{green}{Improved} and \textcolor{red}{deteriorated} pixels are highlighted.}
    \label{fig:driving:cityscapes_evol}
\end{figure*}

\textbf{Visualization on test datasets}
This part presents the visualizations of the \ac{ttaref} \ac{tta} method on the test datasets. The \ac{ttaref} method was selected because among the best performing methods, it is the most novel in the image segmentation setup and has shown particularly strong performance on images most severely impacted by domain shift. The visualizations can be found in Figure \ref{fig:voc_evol} (VOC), Figure \ref{fig:driving:acdc_fog_evol} (ACDC-fog), Figure \ref{fig:driving:acdc_night_evol} (ACDC-night), Figure \ref{fig:driving:acdc_snow_evol} (ACDC-snow), Figure \ref{fig:driving:acdc_rain_evol} (ACDC-rain)
and Figure \ref{fig:driving:cityscapes_evol} (Cityscapes).

\end{document}